\documentclass{article}

\PassOptionsToPackage{numbers, compress}{natbib}

\usepackage[preprint]{neurips_2022}

\usepackage[utf8]{inputenc} 
\usepackage[T1]{fontenc}    
\usepackage{url}            
\usepackage{booktabs}       
\usepackage{amsfonts}       
\usepackage{nicefrac}       
\usepackage{microtype}      
\usepackage{xcolor}         

\usepackage{amsmath,amssymb}
\usepackage{tikz}
\usepackage{comment}
\usepackage{todonotes}
\usepackage{multirow}
\usepackage{color}
\usepackage[colorlinks, linkcolor=red, anchorcolor=blue, citecolor=green]{hyperref}
\usepackage{float}
\usepackage{amsmath,amssymb}
\usepackage{tikz}
\usepackage{comment}
\usepackage{todonotes}
\usepackage{float}
\usepackage{multirow}
\usepackage{color}
\usepackage[colorlinks, linkcolor=red, anchorcolor=blue, citecolor=green]{hyperref}
\usepackage{pifont}
\usepackage{bbding}
\usepackage{wasysym}
\usepackage{makecell}

\title{Rectify ViT Shortcut Learning by Visual Saliency}

\author{%
  Chong Ma \\
  Northwestern Polytechnical University \\
  \texttt{mc-npu@mail.nwpu.edu.cn} \\
  \And
  Lin Zhao \\
  University of Georgia \\
  \texttt{lin.zhao@uga.edu} \\
  \And
  Yuzhong Chen \\
  University of Electronic Science and Technology of China \\
  \texttt{chenyuzhong211@gmail.com} \\
    \And
  David Weizhong Liu \\
  Athens Academy \\
  \texttt{david.weizhong.liu@gmail.com} \\
      \And
  Xi Jiang \\
  University of Electronic Science and Technology of China \\
  \texttt{xijiang@uestc.edu.cn} \\
      \And
  Tuo Zhang \\
  Northwestern Polytechnical University \\
  \texttt{tuozhang@nwpu.edu.cn} \\
      \And
  Xintao Hu \\
  Northwestern Polytechnical University \\
  \texttt{xhu@nwpu.edu.cn} \\
      \And
  Dinggang Shen \\
  ShanghaiTech University \\
  \texttt{dgshen@shanghaitech.edu.cn} \\
      \And
  Dajiang Zhu \\
  University of Texas at Arlington \\
  \texttt{dajiang.zhu@uta.edu} \\
      \And
  Tianming Liu \\
  University of Georgia \\
  \texttt{tianming.liu@gmail.com} \\
}

\begin{document}

\maketitle

\begin{abstract}

  Shortcut learning is common but harmful to deep learning models, leading to degenerated feature representations and consequently jeopardizing the model's generalizability and interpretability. However, shortcut learning in the widely used Vision Transformer (ViT) framework is largely unknown. Meanwhile, introducing domain-specific knowledge is a major approach to rectifying the shortcuts, which are predominated by background related factors. For example, in the medical imaging field, eye-gaze data from radiologists is an effective human visual prior knowledge that has the great potential to guide the deep learning models to focus on meaningful foreground regions of interest. However, obtaining eye-gaze data is time-consuming, labor-intensive and sometimes even not practical. In this work, we propose a novel and effective saliency-guided vision transformer (SGT) model to rectify shortcut learning in ViT with the absence of eye-gaze data. Specifically, a computational visual saliency model (either pre-trained or fine-tuned) is adopted to predict saliency maps for input image samples. Then, the saliency maps are used to distil the most informative image patches. In the proposed SGT, the self-attention among image patches focus only on the distilled informative ones. Considering this distill operation may lead to global information lost, we further introduce, in the last encoder layer, a residual connection that captures the self-attention across all the image patches. The experiment results on four independent public datasets (including two natural and two medical image datasets) show that our SGT framework can effectively learn and leverage human prior knowledge without eye gaze data and achieves much better performance than baselines. Meanwhile, it successfully rectifies the harmful shortcut learning and significantly improves the interpretability of the ViT model, demonstrating the promise of transferring human prior knowledge derived visual saliency in rectifying shortcut learning.

\end{abstract}

\section{Introduction}
As deep learning has been increasingly adopted in natural language processing, computer vision, and medical imaging~\cite{lecun2015deep}, among others, the black box issue that comes with it has attracted extensive attention and discussion.
Especially when deep learning models step into risk-sensitive scenarios such as autonomous driving or medical diagnostics in real-world applications, questions about their interpretability and transferability/generalizability have become more intensive.
Many researchers have come to believe that shortcut learning, as an implicitly harmful phenomenon, is a major reason causing the poor robustness and low transferability/generalizability of deep learning models~\cite{RobertGeirhos2020ShortcutLI} when applying to new scenario or datasets.

Recent work~\cite{XuLuo2022RectifyingTS} revealed that background related factors can be representative and predominant shortcuts in Few-Shot Learning (FSL) that drastically affects the model’s performance. The similar problems also exist in the medical image analysis field, and may cause even more harmful~\cite{zech2018variable, luo2021rethinking, robinson2021deep} outcomes when training deep models for clinical diagnosis. Among different strategies aiming to address shortcut learning, domain-specific knowledge is considered an important approach to avoiding the unintended representations/features learned by shortcut learning. That is, making the model pay more attention to the important regions in the image, rather than being distracted by irrelevant information (e.g., background). For example, in natural images ~\cite{YaoRong2022HumanAI}, eye gaze data can be used to guide fine-grained image classification. In medical images ~\cite{wang2022follow}, radiologists' eye gaze data can be particularly helpful in assisting the disease diagnosis. Eye gaze from radiologists when examining the medical images, as a direct expression of human subjective knowledge or preference, can capture the subtle gaze behavior and therefore identify the areas that are potentially more informative and related to the medical diagnosis tasks. Nevertheless, obtaining this domain/expert knowledge can be time-consuming, labor-intensive and can be even not available in some situations. Thus, an effective way to adapt and leverage the existing data is highly desired. Meanwhile, Vision Transformer (ViT)~\cite{dosovitskiy2020image} has been receiving increasing attention from the computer vision community~\cite{han2022survey} recently. Many variants of the original ViT models have been proposed to improve the performance and efficiency in specific computer vision tasks. In addition, ViT-based models have been exponentially expanded to the applications in medical images, as well as many others. However, there is little research on the shortcut learning issues in the ViT framework.

\begin{figure}
  \centering
  \includegraphics[width=0.8\linewidth]{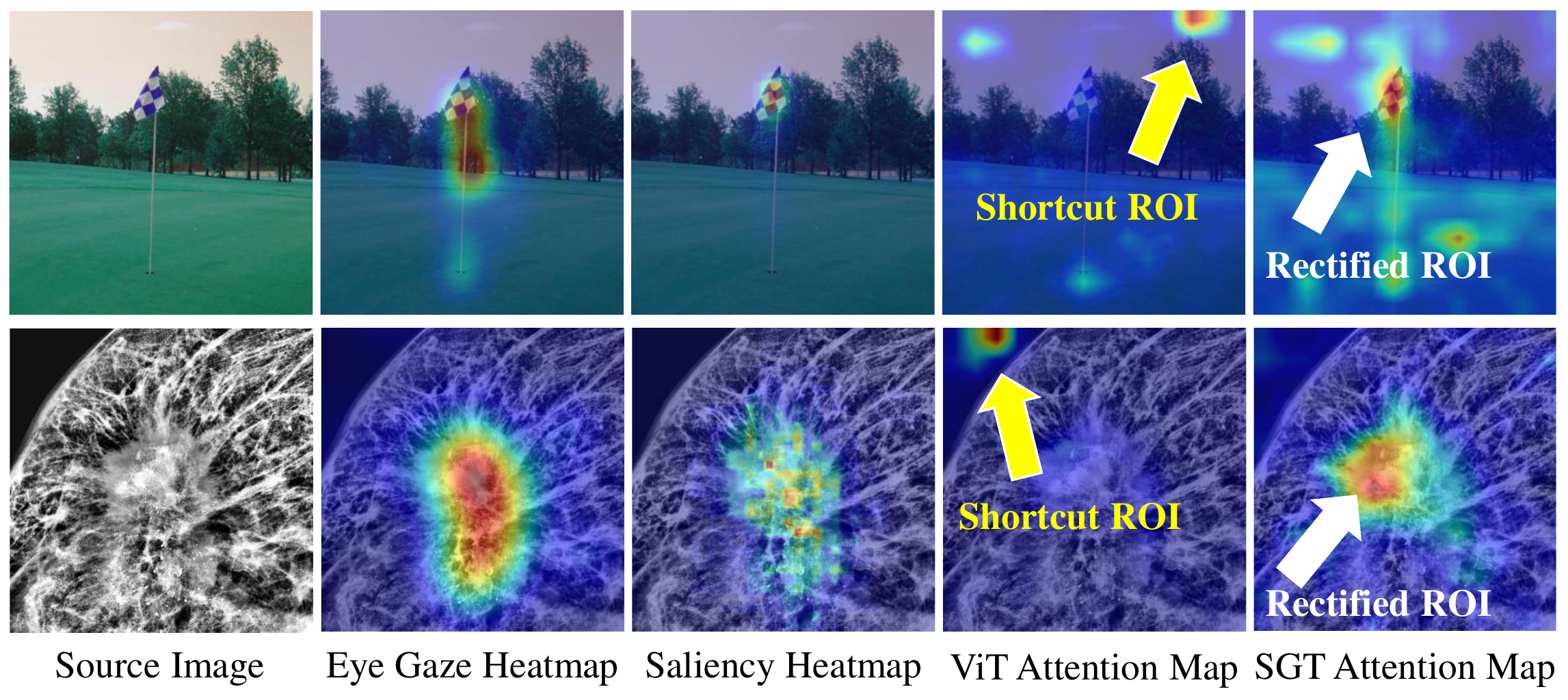}
  \caption{Examples of comparing the predicted saliency map (third column) with eye gaze heatmap (second column), and examples of comparing model's attention map before (fourth column) and after (fifth column) using our saliency guidance method on natural image dataset (first row) and medical image dataset (second row). From the first three columns we can see that the saliency model learns human prior knowledge very well. In the fourth column we can see that without the guidance of prior knowledge, the model can easily focus on shortcut features (as shown by the \textcolor{yellow}{yellow} arrow). In the fifth column, the rectified ROI (regions of interests) of our SGT model focus on more discriminative regions (as shown by the white arrow).}
  \label{fig1}
\end{figure}

In response, in this work, we reveal that ViT models have serious shortcut learning problems, and we propose a novel and effective saliency-guided vision transformer (SGT) model to alleviate the influence of shortcut learning when the domain knowledge such as eye gaze data is not available. Specifically, our SGT firstly learns a saliency map from natural or medical image datasets with eye gaze data, that is, we use real eye gaze data to fine-tune a saliency prediction model to predict the saliency map of images based on human prior knowledge. Then, SGT uses a mask generated from the saliency map to distil the most informative regions from the data without eye gaze information. Our saliency prediction model is trained on both natural and medical image datasets with eye gaze data, for the sake of generating valid saliency maps even on datasets without eye gaze information. Moreover, in order to maintain the global information of the image, we add the residual operation at the last encoder layer of ViT. In this way, our SGT model can learn more effective features from prior visual information and significantly reduce the shortcut learning phenomenon due to dataset size or bias problems, as shown in Fig.~\ref{fig1}. We evaluate our SGT model on four independent public datasets (including two natural and two medical image datasets): FIGRIM~\cite{ZoyaBylinskii2015IntrinsicAE} and CAT2000~\cite{AliBorji2015CAT2000AL} in natural images,  INbreast~\cite{InsMoreira2012INbreastTA} and SIIM-ACR (the dataset of 2019 Pneumothorax Segmentation Challenge~\cite{SIIM-ACR}) in medical images.
Our experiments show that the proposed SGT model significantly improves image classification and medical diagnosis accuracy, and successfully avoids the harmful shortcut learning phenomenon, compared with the baseline ViT models and other popular methods.

Overall, the main contributions this work are:
\begin{itemize}
\item We reveal that the shortcut learning phenomenon is prevalent in the ViT framework, and quantitatively evaluate the severity of shortcuts in some representative ViT baseline models.

\item We propose a novel SGT model, which uses a visual saliency mask operation to guide the model to focus on the regions with relevant and discriminative features.

\item In order to maintain the global information, we add an addition residual connection in our SGT model.
We can apply this strategy to various ViT models, which improves the interpretability and generalizability of the models.


\end{itemize}

\section{Related work}

\paragraph{Shortcut learning}
Recently, shortcut learning has received much attention in deep learning areas such as computer vision (CV)~\cite{CorentinDancette2021BeyondQB, CalebRobinson2021DeepLM, XuLuo2022RectifyingTS} and natural language processing (NLP)~\cite{MengnanDu2021TowardsIA, mccoy2019right, niven2019probing}.
For most of the tasks in deep learning, both the training and test sets come from the same dataset.
This results in some implicit biased information (like cows often appear with the grass, birds often appear with the sky) in the database that exists in both the test and training set.
So the model may learn the background feature instead of the real object, and this kind of shortcut is also hard to find through testing metrics~\cite{RobertGeirhos2020ShortcutLI}.
Many methods have been devised to mitigate the negative effects of shortcuts.
For example,~\cite{XuLuo2022RectifyingTS} proposed a framework named COSOC to extract the foreground objects in images without any extra supervision to avoid harmful background shortcuts in Few-Shot Learning (FSL). 
\cite{JoshuaRobinson2021CanCL} found that the shortcut present in the contrastive learning task suppresses important feature learning. 
They modified the positive and negative samples by implicit feature modification (IFM) to mitigate this phenomenon.
In the medical imaging field, \cite{CalebRobinson2021DeepLM} used feature disentanglement in a multi-task training framework to prevent shortcut learning in the context of automated classification of COVID-19 chest X-ray images.
In contrast to the above approach, we believe that introducing human prior knowledge and using it to guide the deep learning network to focus on more important and relevant regions can be effective in avoiding the shortcut learning phenomenon.

\paragraph{Human eye gaze}
Eye gaze data can be directly obtained by using eye tracking methods~\cite{AndrewTDuchowski2003EyeTM} and showing the interest area by human in image or other stimulus.
In computer vision, many works~\cite{NourKaressli2016GazeEF, QiuxiaLai2021UnderstandingMA, AnthonySantella2006GazebasedIF, YaoRong2022HumanAI} used eye gaze data directly as auxiliary information to perform tasks such as image classification, video action recognition or image segmentation.
For example, ~\cite{QiuxiaLai2021UnderstandingMA} used eye gaze data to accomplish salient target segmentation, video action classification, and fine-grained image classification, respectively.
In the literature, many works~\cite{JosephNStember2019EyeTF, SuneetaMall2018ModelingVS, SuneetaMall2019MissedCA, wang2022follow} have demonstrated the important role of eye gaze data in medical image analysis as well.
For instance, ~\cite{wang2022follow} used the radiologists' eye gaze data to generate the attention map and used it to constrain the model's attention.
\cite{JosephNStember2019EyeTF} developed a CNN-based segmentation method using the information from the radiologists' gaze on the lesion.
However, collecting human eye gaze data is time-consuming and laborious, especially for collecting data from clinical radiologists.

\paragraph{Visual saliency prediction}
An effective alternative way is to use saliency model to predict saliency maps that attract human's visual attention.
Nowadays, there are two main tasks in artificial saliency: saliency prediction and salient object detection.
The saliency prediction model outputs a Gaussian distributed heatmap for direct simulation of human fixation information.
For example, many works~\cite{XunHuang2015SALICONRT, MarcellaCornia2016ADM, MatthiasKmmerer2017UnderstandingLA, SenJia2018EMLNETAnEM} use CNN-based models to generate the saliency map of images.
Among them, EML-NET~\cite{SenJia2018EMLNETAnEM} performs well on several benchmarks~\cite{AliBorji2021SaliencyPI}.
In comparison, the salient object detection (SOD) model outputs a mask map, which simulates human visual perception in locating the most significant object(s) in a scene.
Many works such as~\cite{WenguanWang2018SalientOD, JiangJiangLiu2019ASP, JiaxingZhao2019EGNetEG, ZheWu2019StackedCR, KerenFu2020JLDCFJL} used fully convolutional network (FCN)-based or VGG-based model to generate the salient object map.
Recently, the VST~\cite{NianLiu2021VisualST} model achieves excellent prediction results on both RGB and RGB-D datasets based on the transformer architecture.
In order to be independent on the eye gaze data and to obtain the mask discussed in Sec.~\ref{saliency_mask}, we use saliency prediction model, which outputs a heatmap with a Gaussian distribution that is more convenient for mask generation.
With this model, we can obtain valuable information without human involvement or intervention. 
Especially within the medical imaging field, a capable saliency prediction model can simulate a radiologist expert to identify areas of lesions or interests in medical images.

\paragraph{Vision Transformer}
Since Vision Transformer (ViT)~\cite{dosovitskiy2020image} has been introduced in the field of computer vision, more and more self-attention based approaches have been proposed. 
DeiT~\cite{HugoTouvron2020TrainingDI} introduced a distillation strategy that allowed ViT models to be trained more efficiently on smaller datasets.
\cite{XiangxiangChu2021DoWR} proposed a Position Encoding Generator (PEG) that can deal with the input sequences of arbitrary length.
Swin Transformer~\cite{ZeLiu2021SwinTH} developed a hierarchical feature representation and shifted window operation.
Afterwards, researchers proposed Swin Transformer V2~\cite{ZeLiu2022SwinTV} with higher parameters and the ability to handle larger image size ($1536 \times 1536$). 
However, in many cases, we do not have sufficiently large data to train a good ViT model. In order to improve the performance of ViT models with small-scale data, we introduce image saliency as a prior knowledge to guide the ViT models to focus on more relevant and discriminative features.

\section{Method}
In this section, we firstly illustrate the design of SGT in Sec.~\ref{model}. 
Then we introduce the pre-processing of saliency map in Sec.~\ref{saliency_model} and the generation of saliency mask in Sec.~\ref{saliency_mask}.

\subsection{Saliency-guided Vision Transformer}
\label{model}

\begin{figure}
  \centering
  \includegraphics[width=1.0\linewidth]{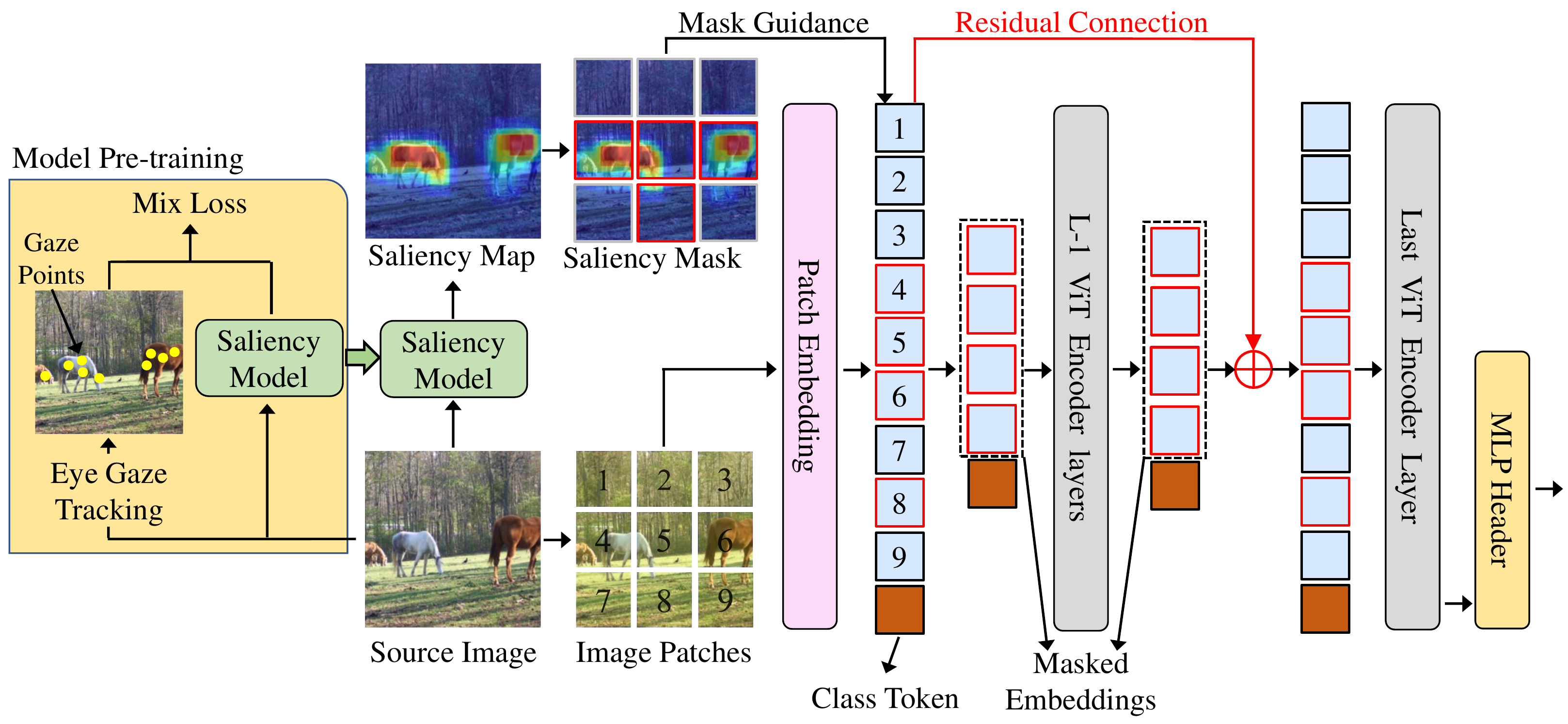}
  \caption{The architecture of our proposed SGT model. First we use the eye gaze data (\textcolor{yellow}{yellow} points) collected on the image as ground truth to train a saliency model to be used in place of humans. Then the saliency map is predicted by our saliency model and processed to generate saliency mask. Next, we applied the mask to the embedding features after patch embedding to filter out the most important features (highlighted by \textcolor{red}{red} rectangular). The masked embedding features are treated as input to the transformer encoder layers. After L-1 layers of transformer encoder, we applied residual connection before the last encoder layer (highlighted by \textcolor{red}{red} arrow). Finally, we use the class token as the representation features to perform the classification task.}
  \label{fig2}
\end{figure}

In this work, to avoid the ViT model from learning irrelevant or harmful shortcuts, we use visual saliency to guide the model to focus on the most important and relevant regions.
We implemented this idea by introducing a saliency mask on input image patches and an additional residual connection on the SGT model.
Fig.~\ref{fig2} shows the overall architecture of our proposed SGT model. 
In the vanilla ViT framework, an image is divided into multiple patches after patch embedding, and these patches with a class token are input to the transformer encoder for self-attention operation. 
In order to focus on important image regions, we directly filter out patches with high saliency.
Specifically, the input image can be divided into $N$ patches and the ViT model maps the images patches $x_p^i$ ($i =1,2,\cdots,$N) to \emph{D} dimension embedding with a trainable linear projection $E \in \mathbb{R}^{HWC \times D}$ where $H,W,C$ is the height, weight and channel of the images.
After patch embedding, we directly use the mask generated by saliency map to filter out the most important and relevant patches and obtain $\tilde{z}_0$:

\begin{equation}
\label{eq_mask_patch}
    \tilde{z}_0=[x_{class};(x_p^1E;x_p^2E;\cdots;x_p^NE) \odot Mask]+E_{pos}
\end{equation}

where $x_{class} \in \mathbb{R}^N$ is the class token for classification, $E_{pos} \in \mathbb{R}^{(N+1) \times D}$ is the learnable position embedding, $(x_p^1E;x_p^2E;\cdots;x_p^NE)$ are the original patch embeddings and $Mask\in\mathbb{R}^N$ is the binary saliency mask detailed in Sec.~\ref{saliency_mask}.
Then we obtain the important patch embeddings $\tilde{z}_0$ and input to the first transformer encoder layer.


For Vision Transformer \cite{dosovitskiy2020image}, the forward propagation of each transformer encoder layer can be written as:

\begin{gather}
        \tilde{z}_l^\prime = MSA(LN(\tilde{z}_{l-1}))+\tilde{z}_{l-1}\\
         \tilde{z}_l = MLP(LN(\tilde{z}_l^\prime))+\tilde{z}_l^\prime
\end{gather}

where $\tilde{z}_l^\prime$ is the \emph{l}-th layer's masked embedding patches.
$MSA$, $MLP$ and $LN$ are the multiheaded self-attention, multilayer perceptron, and layer norm in each block.

However, masking some patches in the first layer results in a risk of missing useful background information and positional relationships among blocks.
Inspired by \cite{He_2016_CVPR,he2021masked}, we added the whole initial embedding patches back to the last layer's embedding patches to retain global information and maintain the correlations of all patches.
Therefore, the input embedding patch of last transformer encoder $\hat{z}_{l-1}$ can be written as:

\begin{equation}
    \hat{z}_{l-1}=[x_{class}; z_0^{1:N} + \tilde{z}_{l-1}]
\end{equation}

where $z_0^{1:N} + \tilde{z}_{l-1}$ means that $\tilde{z}_{l-1}$ is added to $z_0^{1:N}$ according to the corresponding position of mask.

\subsection{Saliency heatmap generation}
\label{saliency_model}
We use a computational visual saliency model EML-NET~\cite{SenJia2018EMLNETAnEM} to predict saliency heatmap.
In brief, EML-NET is a scalable encoder-decoder framework that leverages multiple deep CNN models (such as VGG-16~\cite{KarenSimonyan2015VeryDC}, ResNet-50~\cite{He_2016_CVPR}, DenseNet-161~\cite{GaoHuang2016DenselyCC} and NasNetLarge~\cite{BarretZoph2017LearningTA}) to better extract visual features for saliency prediction.
And it uses a combined loss function consisting three saliency metrics including the Kullback-Leibler Divergence (KLD)~\cite{ZoyaBylinskii2016WhatDD}, the modified Pearson's Correlation Coefficient~\cite{SenJia2018EMLNETAnEM} and Normalized Scanpath Saliency~\cite{SenJia2018EMLNETAnEM}.
In our study, due to that EML-NET was trained using the SALICON dataset ~\cite{MingJiang2015SALICONSI}, the largest dataset for saliency prediction in natural images, we directly use the pre-trained EML-NET to generate saliency maps for the natural image datasets (FIGRIM~\cite{ZoyaBylinskii2015IntrinsicAE} and CAT2000~\cite{AliBorji2015CAT2000AL}). 
For the medical image datasets (INbreast~\cite{InsMoreira2012INbreastTA} and SIIM-ACR~\cite{SIIM-ACR}), the pre-trained EML-NET is fine-tuned using the accompanied eye-tracking recording data.

In this way, we can generate effective saliency maps for both natural and medical images regardless of whether the image has corresponding eye gaze data or not.
As a comparison study, we also compared the proposed SGT model with the saliency maps generated by three different ways, that is, the one predicted by the EML-NET, the one predicted by a SOD model~\cite{NianLiu2021VisualST} and the one generated directly using the eye-gaze recordings.
All comparison results are shown in Sec.~\ref{ablation}.

\subsection{Saliency mask generation}
\label{saliency_mask}

\begin{figure}[htb]
  \centering
  \includegraphics[width=0.9\linewidth]{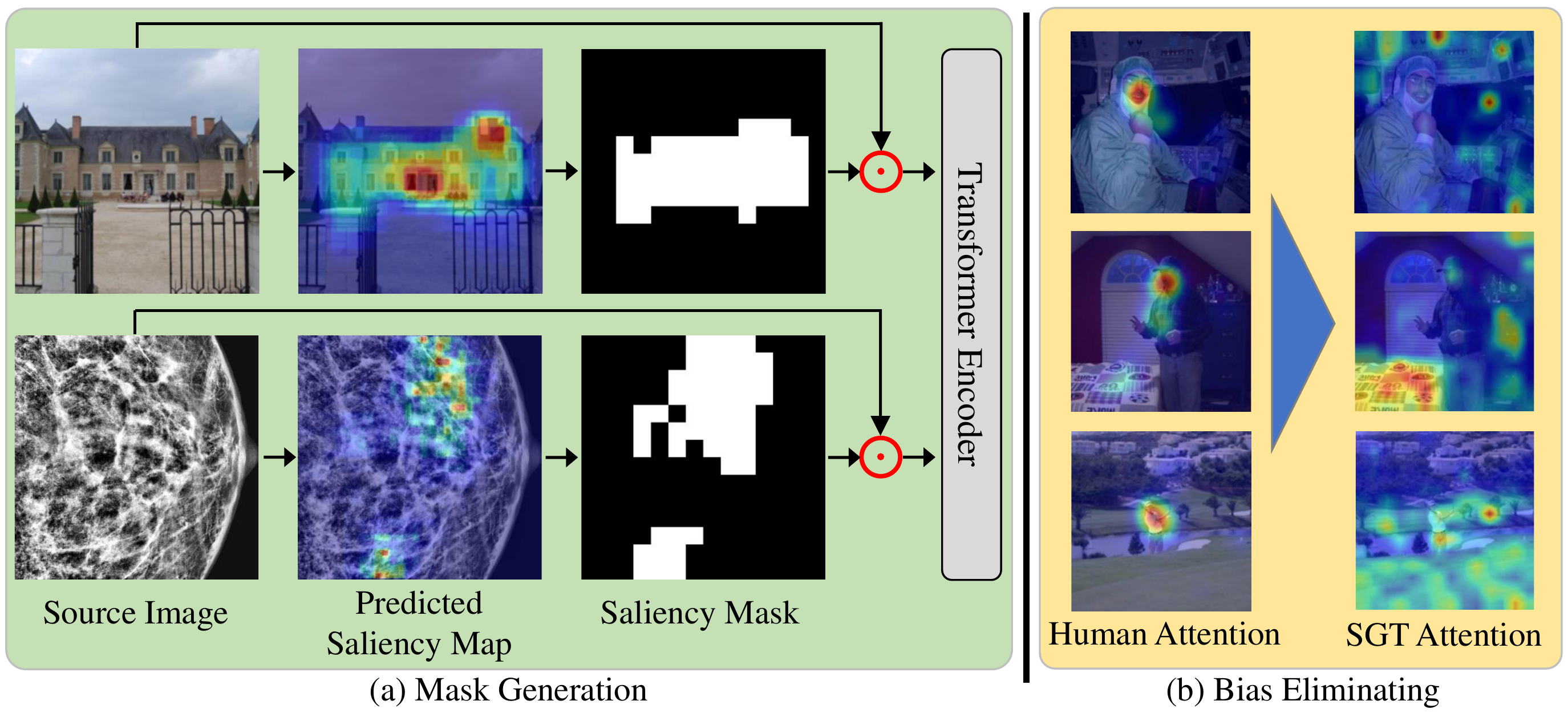}
  \caption{(a) The generation of our saliency mask in natural (first row) and medical (second row) datasets. (b) Elimination of human attention bias.}
  \label{fig3}
\end{figure}

In this work, the saliency mask is used to filter out the most important and relevant regions of the image for the model training.
As shown in Fig.~\ref{fig3}(a), we first use the saliency model introduced in Sec.~\ref{saliency_model} to predict the saliency map of the images.
As the patch embedding layer in ViT model maps the input image to $14 \times 14 \times \emph{C}$ (suppose the image size is $224 \times 224$ and the patch size is $16 \times 16$), we also resize the saliency map to $14 \times 14$.
Then the binary saliency mask is obtained by setting the largest $M$ values to 1 and 0 to the rest on saliency map, where $M$ is set to 49 ($7 \times 7$), 64 ($8 \times 8$) or 81 ($9 \times 9$) in this work, meaning that we mask out about 75\%, 70\% or 60\% of the unimportant areas of the image.
Finally, we multiply the saliency mask by the features after patch embedding and thus input them to the subsequent transformer encoder layers.

However, human gaze information is not always effective for the classification task, especially in aimless observing tasks.
As shown in Fig.~\ref{fig3}(b), people focus on the characters in the image instead of the scene information.
It means that our predicted saliency map would have a similar bias.
Therefore, we introduce a random guidance strategy to randomly adopt the saliency mask during the training of our SGT model, and after experimental verification (see Appdx. A) we set a threshold value of 0.5.
It allows for a balance of prior knowledge guidance and independent learning from the model.
As Fig.~\ref{fig3}(b) shows, this strategy diminishes the impact of these human misguided biases.

\section{Experiments}

\subsection{Datasets and evaluation metrics}
\label{datasets}
We evaluate our SGT model on four different datasets, including two natural image datasets FIGRIM~\cite{ZoyaBylinskii2015IntrinsicAE} and CAT2000~\cite{AliBorji2015CAT2000AL}, two medical image datasets INbreast~\cite{InsMoreira2012INbreastTA} and SIIM-ACR~\cite{SIIM-ACR}.
Among them, FIGRIM~\cite{ZoyaBylinskii2015IntrinsicAE} provides eye fixation data for a total of 2787 images spanning 21 indoor and outdoor scene categories.
CAT2000~\cite{AliBorji2015CAT2000AL} has 2000 training images and 2000 testing images from 20 different categories with eye tracking data.
The SIIM-ACR~\cite{SIIM-ACR} is a chest X-ray dataset with only pneumothorax disease, and recently work~\cite{KhaledSaab2021ObservationalSF} randomly selected 1,170 images, with 268 cases of pneumothorax and collected gaze data from three radiologists.
Follow this work~\cite{KhaledSaab2021ObservationalSF}, we chose these 1170 images and randomly split them into 80\% and 20\% as training and testing dataset.
The INbreast~\cite{InsMoreira2012INbreastTA} dataset includes 410 full-ﬁeld digital mammography images which were collected during low-dose X-ray irradiation of the breast.
Also, we invited a radiologist with 10 years of experience of diagnosing cancers and collected the eye gaze data using an eye-tracking system.
To augment and balance the data, we randomly cropped the original images and divided them into normal, benign and malignant categories. 
More details of the data pre-processing can be found in Appdx. D.

We adopt accuracy (Acc), F1-score (F1) and area under curve (AUC) evaluation metrics to evaluate our model performance comprehensively.
To check whether there is shortcut learning phenomenon, we use the Grad-CAM~\cite{RamprasaathRSelvaraju2016GradCAMVE} method to visualize the heatmap of models on the image and we only consider whether the interest region of model has a significant relationship with the category of image.
For example, if the model prediction is "house", but it pays attention to the sky or the grass, we consider that the model learns the shortcut feature.
We counted these samples in each experiment to eventually obtain the proportion of shortcut learning (PSL) in the dataset. More details of PSL can be found in Appdx. E.

\subsection{Implementation details}
\label{implementation}
For all experiments, we trained 60 epochs with 0.0001 initial learning rate and choose a cosine decay learning rate scheduler and 5 epochs of warm-up with an Adam optimizer.
All saliency maps were extracted by our saliency prediction model, and all images were simply resized to $224 \times 224$ pixels.
More implementation details can be found in Appdx. D.

\subsection{Comparison with ViT baselines}

\begin{figure}[htb]
  \centering
  \includegraphics[width=1.0\linewidth]{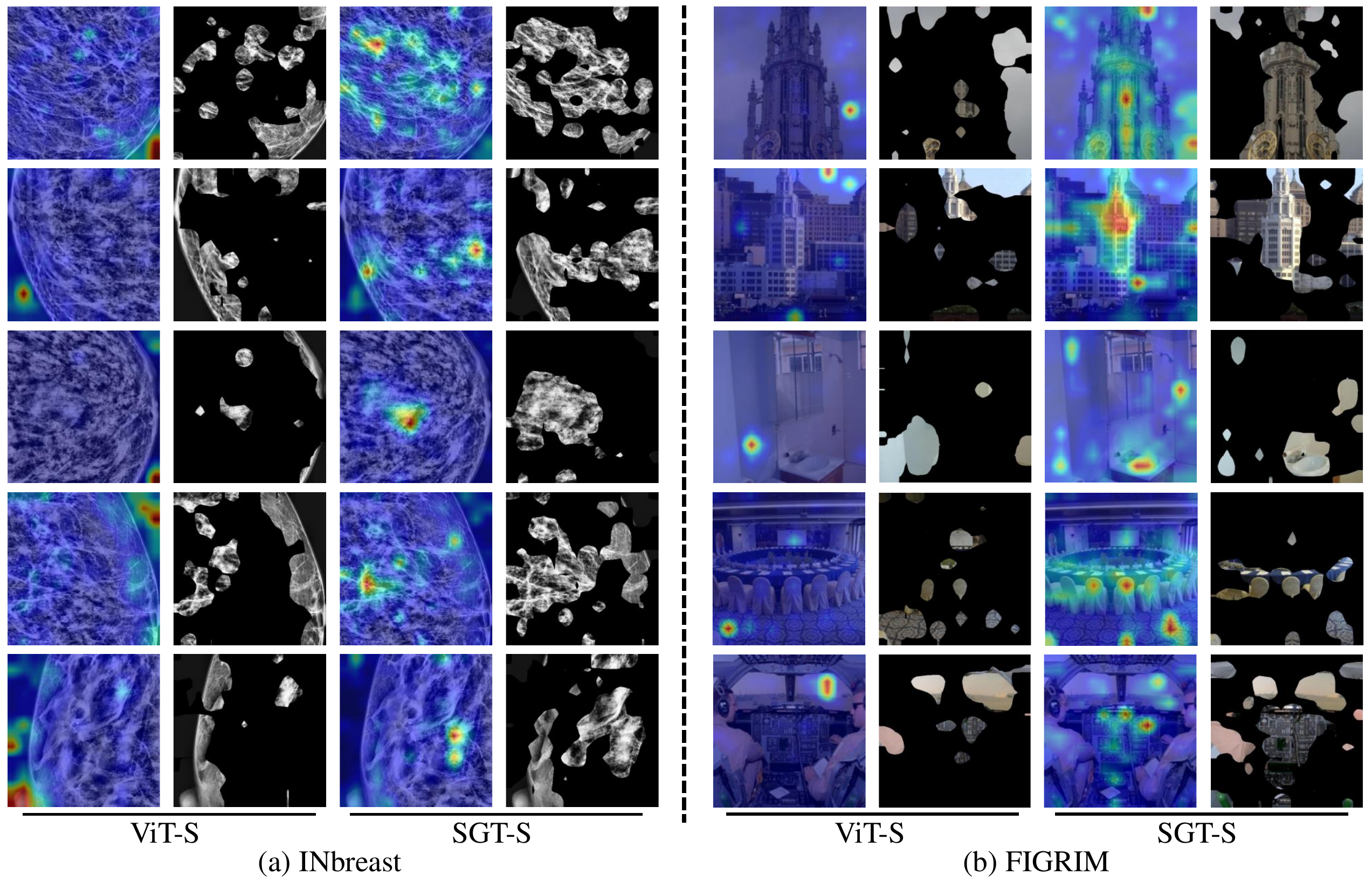}
  \caption{Comparison examples on INbreast (a) and FIGRIM (b) dataset. We choose ViT-S and SGT-S model to demonstrate the improvement of our proposed method for the shortcut learning phenomenon. For each model, we used a pair of figures to show the attention of the model. The left side is the attention map of the model obtained by Grad-CAM, and the right side is the image activation region. Each row corresponds to the same example.}
  \label{fig4}
\end{figure}

In this paper, ViT-S, ViT-B and ViT-L represent the Small, Base and Large scales of the ViT baseline, respectively, and SGT-S, SGT-B and SGT-L represent the application of our SGT model on the corresponding three scales.
All scales of ViT and SGT models were pre-trained on ImageNet dataset~\cite{deng2009imagenet} and fine-tuned on each dataset in our experiment.
Fig.~\ref{fig4} shows some comparison examples on INbreast~\cite{InsMoreira2012INbreastTA} and FIGRIM~\cite{ZoyaBylinskii2015IntrinsicAE} dataset. 
In Fig.~\ref{fig4}(a), the ViT-S model without the guidance of saliency tends to focus on the background region in mammogram images, as the big difference between breast tissue and background.
We also found that the ViT-S model often incorrectly focuses on the nipple region, as shown in the last row of Fig.~\ref{fig4}(a).
This is because there are no nipple-related diseases in INbreast~\cite{InsMoreira2012INbreastTA} dataset and the model focuses on it only because of its characteristic appearance.
In contrast, our proposed SGT-S model corrected this shortcut learning phenomenon.
As we can see in the first two rows of Fig.~\ref{fig4}(b), since the "skyscraper" often appear together with sky, the ViT-S model tends to focus on the more distinguishable sky areas.
With our proposed saliency guidance, the SGT-S model also corrected this shortcut learning phenomenon.

Tab.~\ref{table1} reports the comparison results of ViT and our SGT model at different scales on four datasets.
The percentage of shortcut learning (PSL) of all experiments are significantly decreased after using our saliency guidance method.
For the other three metrics, our SGT model is mostly better than the ViT baselines.
In particular, for FIGRIM~\cite{ZoyaBylinskii2015IntrinsicAE} dataset, although metrics of the ViT-S and ViT-L are better than SGT, this advantage is brought by learning shortcut features.
The interesting phenomenon here, however, is that although our SGT method suppresses the shortcut learning, there is also a slight decrease in classification performance.
This experiment shows that the shortcut learning phenomenon is common in ViT models, especially for simpler models and small datasets.

\setlength{\tabcolsep}{4pt}
\begin{table}
\begin{center}
\caption{Comparison results between ViT baselines and our proposed SGT model on four datasets in natural and medical images. The Accuracy, AUC, F1-score and the percentage of shortcut learning (PSL) are reported. The~\textbf{bold} numbers denote the better comparison result.}
\label{table1}
\begin{tabular}{cccccccc}
\hline\noalign{\smallskip}
Dataset &Metrics &ViT-S~\cite{dosovitskiy2020image} &SGT-S &ViT-B~\cite{dosovitskiy2020image} &SGT-B &ViT-L~\cite{dosovitskiy2020image} &SGT-L\\
\noalign{\smallskip}
\hline
\noalign{\smallskip}
INbreast~\cite{InsMoreira2012INbreastTA}
    &Acc. $\uparrow$   &91.57  &\textbf{93.98}  &90.36  &\textbf{91.57}  &90.36  &\textbf{92.77} \\
    &AUC $\uparrow$    &88.79  &\textbf{91.55}  &90.18  &\textbf{92.10}  &94.94  &\textbf{96.91} \\
    &F1 $\uparrow$     &91.01  &\textbf{94.07}  &90.45  &\textbf{91.95}  &90.30  &\textbf{92.71} \\
    &PSL $\downarrow$    &50.00  &\textbf{38.29}  &42.20  &\textbf{37.10}  &43.90  &\textbf{19.00} \\
\hline
\noalign{\smallskip}
SIIM-ACR~\cite{SIIM-ACR}        
    &Acc. $\uparrow$   &83.20  &\textbf{86.00}  &85.20  &\textbf{86.80}  &\textbf{86.80}  &85.60 \\
    &AUC $\uparrow$    &77.53  &\textbf{85.53}  &81.50  &\textbf{85.46}  &84.20  &\textbf{86.61} \\
    &F1 $\uparrow$     &82.67  &\textbf{85.36}  &84.92  &\textbf{86.32}  &\textbf{86.32}  &84.57 \\
    &PSL $\downarrow$     &47.60  &\textbf{43.20}  &51.60  &\textbf{42.00}  &58.80  &\textbf{45.60} \\
\hline
\noalign{\smallskip}
FIGRIM~\cite{ZoyaBylinskii2015IntrinsicAE}
    &Acc. $\uparrow$   &\textbf{95.13}  &94.81  &94.97  &\textbf{95.28}  &\textbf{95.60}  &95.13 \\
    &AUC $\uparrow$    &\textbf{99.47}  &99.26  &\textbf{99.77}  &99.72  &\textbf{99.90}  &99.81 \\
    &F1 $\uparrow$     &\textbf{95.12}  &94.84  &94.94  &\textbf{95.24}  &\textbf{95.53}  &95.09 \\
    &PSL $\downarrow$     &43.40  &\textbf{35.06}  &39.47  &\textbf{35.85}  &55.50  &\textbf{48.90} \\
\hline
\noalign{\smallskip}
CAT2000~\cite{AliBorji2015CAT2000AL}
    &Acc. $\uparrow$   &85.47  &\textbf{86.87}  &86.33  &\textbf{87.40}  &87.87  &\textbf{89.27} \\
    &AUC $\uparrow$    &98.39  &\textbf{98.57}  &98.89  &\textbf{98.95}  &\textbf{98.95}  &98.83 \\
    &F1 $\uparrow$     &84.94  &\textbf{86.59}  &85.79  &\textbf{87.26}  &87.65  &\textbf{89.09} \\
    &PSL $\downarrow$     &30.13  &\textbf{23.13}  &26.07  &\textbf{14.20}  &31.47  &\textbf{16.80} \\

\hline
\end{tabular}
\end{center}
\end{table}
\setlength{\tabcolsep}{1.4pt}

\subsection{Comparison with popular models}

In this subsection, we compare our SGT model with other popular models on INbreast~\cite{InsMoreira2012INbreastTA} and FIGRIM~\cite{ZoyaBylinskii2015IntrinsicAE} dataset, such as the family of ResNet~\cite{He_2016_CVPR}, Swin Transformer V1~\cite{ZeLiu2021SwinTH}, EfficientNet-B0~\cite{tan2019efficientnet} and EfficientNet-B7~\cite{tan2019efficientnet}.
As shown in Tab.~\ref{table2}, our proposed SGT model significantly outperforms the other methods in Accuracy, AUC and F1 score.
In particular, the PSL metric has a clear advantage on INbreast~\cite{InsMoreira2012INbreastTA}, but is far inferior to ResNet~\cite{He_2016_CVPR} on FIGRIM~\cite{ZoyaBylinskii2015IntrinsicAE}.
We believe that there are two main reasons for this.
One is that eye gaze data in medical images come from radiologists and are highly task-driven, whereas eye gaze data in natural images come from the aimless gaze behavior of ordinary people and are far less instructive than those of radiologists.
Another reason is that the inductive bias inherent in CNN models is more consistent with the human visual system, and the models are pre-trained on large-scale natural image datasets, so the area concerned by CNN models is also more related to human attention than ViT models on FIGRIM~\cite{ZoyaBylinskii2015IntrinsicAE}.
It can be seen from the Tab.~\ref{table2} that the shortcut in medical images has a much greater effect in natural images.

\setlength{\tabcolsep}{4pt}
\begin{table}
\begin{center}
\caption{Comparison results with other popular methods on INbreast and FIGRIM dataset. The Accuracy, AUC, F1-score and the percentage of shortcut learning (PSL) are reported. \textcolor{red}{Red} and \textcolor{blue}{blue} denote the best and the second-best results, respectively.}
\label{table2}
\begin{tabular}{ccccccccc}
\hline\noalign{\smallskip}
\multirow{2}{*}{Method}&\multicolumn{4}{c}{INbreast~\cite{InsMoreira2012INbreastTA}} & \multicolumn{4}{c}{FIGRIM~\cite{ZoyaBylinskii2015IntrinsicAE}}\\ \cmidrule(r){2-5} \cmidrule(r){6-9}
 &Acc. $\uparrow$ & AUC $\uparrow$ & F1 $\uparrow$ & PSL $\downarrow$  & Acc. $\uparrow$ & AUC $\uparrow$ & F1 $\uparrow$ &PSL $\downarrow$ \\
\noalign{\smallskip}
\hline
\noalign{\smallskip}
ResNet-18~\cite{He_2016_CVPR}         
&87.95      &94.86      &88.55      &52.44      &86.95      &99.44         &86.88      &17.92  \\
ResNet-50~\cite{He_2016_CVPR}        
&89.16      &\textcolor{blue}{96.55}      &89.35      &41.71      &89.78      &99.57         &89.55      &\textcolor{blue}{15.09}  \\
ResNet-101~\cite{He_2016_CVPR}        
&90.36      &96.00      &90.71      &37.80      &90.09      &99.57         &90.09      &\textcolor{red}{11.79}  \\
EfficientNet-B0~\cite{tan2019efficientnet}
&86.75  &88.84  &87.20  &46.85          &84.28  &98.64     &84.18  &29.69  \\
EfficientNet-B7~\cite{tan2019efficientnet}
&91.56  &92.08  &91.64  &38.52          &90.72  &99.47     &90.66  &26.83  \\
SwinT V1~\cite{ZeLiu2021SwinTH}       
&91.57      &88.84      &91.64      &51.71          &92.77      &99.32         &92.75      &49.69  \\
SGT-S (ours)                       
&\textcolor{red}{93.98}  &91.55  &\textcolor{red}{94.07}  &38.29      &\textcolor{blue}{94.81}  &99.26   &\textcolor{blue}{94.84}    &35.06  \\
SGT-B (ours)                      
&\textcolor{blue}{91.57}  &92.10  &\textcolor{blue}{91.95}  &\textcolor{blue}{37.10}      &\textcolor{red}{95.28}  &\textcolor{blue}{99.72}   &\textcolor{red}{95.24}    &35.85  \\
SGT-L (ours)                      
&92.77  &\textcolor{red}{96.91}  &92.71  &\textcolor{red}{19.00}    &95.13 &\textcolor{red}{99.81}   &95.09    &48.90  \\

\hline
\end{tabular}
\end{center}
\end{table}
\setlength{\tabcolsep}{1.4pt}

\subsection{Ablation study}
\label{ablation}
We considered the effect of different types of masks and conducted comparative experiments on INbreast~\cite{InsMoreira2012INbreastTA} and FIGRIM~\cite{ZoyaBylinskii2015IntrinsicAE} dataset using SGT-S model.
Each row of the mask type in Tab.~\ref{table3} represents (from top to bottom): without guidance, human attention guidance, guidance of random position mask, guidance mask generated by Grad-CAM~\cite{RamprasaathRSelvaraju2016GradCAMVE} of the model, guidance mask generated by a SOD model VST~\cite{NianLiu2021VisualST}, and the last three rows use different scales of our predicted saliency mask.
As shown in Tab.~\ref{table3}, the performance of directly using eye gaze is significantly better, with our SGT model being the next best in INbreast~\cite{InsMoreira2012INbreastTA}.
In FIGRIM~\cite{ZoyaBylinskii2015IntrinsicAE}, consistent with Tab.~\ref{table1}, although our model reduces the shortcut learning phenomenon, the other metrics do not improve much and even perform lower than the type without mask.
We believe that this is due to the confusion of category-related objects in FIGRIM~\cite{ZoyaBylinskii2015IntrinsicAE} dataset, which leads the model to learn some shortcut features more easily, instead of the correct and meaningful ones, for the purpose of classification (detailed demonstration can be found in Appdx. B).
In the comparison results of different scales of mask, we found that medical images are more suitable for small, focused regions, while natural images need relatively large ones.
This may be because the area of the lesion in medical images is usually smaller than the object in the natural images.
For more experiment results on the use of mask guidance on different encoder layers of ViT and whether to add a residual connection, see Appdx. C.

\setlength{\tabcolsep}{4pt}
\begin{table}
\begin{center}
\caption{Ablation study. The Accuracy, AUC, F1-score and the percentage of shortcut learning (PSL) are reported. \textcolor{red}{Red} and \textcolor{blue}{blue} denote the best and the second-best results, respectively.}
\label{table3}
\begin{tabular}{ccccccccc}
\hline\noalign{\smallskip}
\multirow{2}{*}{Mask Type}&\multicolumn{4}{c}{INbreast~\cite{InsMoreira2012INbreastTA}} & \multicolumn{4}{c}{FIGRIM~\cite{ZoyaBylinskii2015IntrinsicAE}}\\ \cmidrule(r){2-5} \cmidrule(r){6-9}
 &Acc. $\uparrow$ & AUC $\uparrow$ & F1 $\uparrow$ & PSL $\downarrow$  & Acc. $\uparrow$ & AUC $\uparrow$ & F1 $\uparrow$ &PSL $\downarrow$ \\
\noalign{\smallskip}
\hline
\noalign{\smallskip}
W/O Mask         
&91.35  &88.79  &91.01  &50.00      &\textcolor{blue}{95.12}  &99.47   &\textcolor{blue}{95.12}   &43.40  \\
Eye Gaze        
&\textcolor{red}{93.98}  &\textcolor{red}{94.02}  &93.45  &\textcolor{red}{37.50}      &\textcolor{red}{95.60}  &\textcolor{blue}{99.67}   &\textcolor{red}{95.59}   &39.31  \\
Random Position     
&90.36  &90.18  &89.68  &54.10      &93.24  &99.33   &93.22   &42.45  \\
Grad-CAM~\cite{RamprasaathRSelvaraju2016GradCAMVE}        
&91.57  &92.43  &91.35  &56.80      &94.03  &99.64   &94.06   &39.78  \\
VST~\cite{NianLiu2021VisualST}         
&91.57  &\textcolor{blue}{94.00}  &91.73  &47.56      &93.71  &99.45   &93.68   &36.95  \\
SGT-S (75\% Mask)
&\textcolor{red}{93.98}  &91.55  &\textcolor{red}{94.07}  &\textcolor{blue}{38.29}      &94.50  &\textcolor{red}{99.71}   &94.50   &35.85  \\
SGT-S (70\% Mask)   
&\textcolor{blue}{93.97}  &92.54  &\textcolor{blue}{93.93}  &43.41      &94.81  &99.26   &94.84   &\textcolor{blue}{35.06}\\
SGT-S (60\% Mask)
&91.56  &90.00  &91.01  &55.85      &94.65  &99.6   &94.66   &\textcolor{red}{29.87}  \\

\hline
\end{tabular}
\end{center}
\end{table}
\setlength{\tabcolsep}{1.4pt}

\subsection{Limitation}
\label{limitation}
As mentioned in Sec.~\ref{saliency_mask}, collected eye-gaze data sometimes have a small amount of "human bias", which causes the saliency prediction model to misguide the SGT model.
Although we used a simple strategy to mitigate this phenomenon, how to effectively balance human prior knowledge and model's own learning still needs to be explored.
In addition, the model visualization method used in the experiments has an impact on the statistics of the shortcut learning phenomenon, and better visualization will lead to more accurate results.

\section{Conclusion and discussion}

In this work, we revealed that the shortcut learning phenomenon is prevalent in the ViT framework, and we proposed a novel saliency-guided vision transformer (SGT) to suppress and rectify shortcut learning by infusing artificial prior knowledge.
This artificial prior knowledge comes from the saliency map predicted by a saliency model trained on real human eye gaze data, which allows us to use SGT model on images later without eye gaze data.
We found that sometimes the human eye gaze is not reliable, and there is a balance between the guidance of human prior knowledge and the learning of the model itself.
We also found that the shortcut features present in some datasets, especially small ones, are easier to classify than the correct features.
Therefore, to fully evaluate the performance of the model, multiple test datasets should be used as much as possible to reduce the false positive caused by shortcuts.
We believe that multi-task learning may suppress certain shortcut learning brought by a single task, and in the future, we will do further research based on this and design a more comprehensive evaluation framework for shortcut learning phenomenon.

\bibliographystyle{splncs04}
\bibliography{references}

\appendix

\clearpage

\section{Comparison results of random guidance strategy}
\label{append_a}

To mitigate the possible negative effects of human prior knowledge, we introduce a random guidance strategy to balance prior knowledge with the model's own learning.
Specifically, we first set a threshold value $\emph{T} \in [0, 1]$.
Then, our SGT model generates a random probability $\emph{p} \in [0, 1]$ in each batch of training, and the model uses the saliency mask guidance if $\emph{p} \geq \emph{T}$, and not if $\emph{p} < \emph{T}$.
we selected five thresholds for comparison experiment on the INbreast~\cite{InsMoreira2012INbreastTA} and FIGRIM~\cite{ZoyaBylinskii2015IntrinsicAE} datasets, respectively, based on our SGT-S model.
As shown in Tab.~\ref{append_table1}, we finally chose 0.5 as the threshold in our mask guidance strategy.

\setlength{\tabcolsep}{4pt}
\begin{table}[H]
\begin{center}
\caption{Comparison results of mask guidance with different thresholds. The Accuracy, AUC, F1-score are reported. \textcolor{red}{Red} and \textcolor{blue}{blue} denote the best and the second-best results, respectively.}
\label{append_table1}
\begin{tabular}{ccccccc}
\hline\noalign{\smallskip}
\multirow{2}{*}{Threshold}&\multicolumn{3}{c}{INbreast~\cite{InsMoreira2012INbreastTA}} & \multicolumn{3}{c}{FIGRIM~\cite{ZoyaBylinskii2015IntrinsicAE}}\\ \cmidrule(r){2-4} \cmidrule(r){5-7}
 &Acc. $\uparrow$ & AUC $\uparrow$ & F1 $\uparrow$  & Acc. $\uparrow$ & AUC $\uparrow$ & F1 $\uparrow$  \\
\noalign{\smallskip}
\hline
\noalign{\smallskip}
\emph{T} = 0.2
&90.36  &89.90  &90.67   &94.34   &\textcolor{red}{99.29}  &94.42   \\
\emph{T} = 0.4
&91.57   &90.18   &91.96   &94.03   &99.15   &94.05  \\
\emph{T} = 0.5
&\textcolor{red}{93.98}   &91.55   &\textcolor{red}{94.07}   &\textcolor{red}{94.81}   &\textcolor{blue}{99.26}   &\textcolor{red}{94.84}  \\
\emph{T} = 0.6     
&\textcolor{blue}{93.56}  &\textcolor{red}{92.53}  &\textcolor{blue}{93.68}    &\textcolor{blue}{94.50}  &99.24  &\textcolor{blue}{94.54}  \\
\emph{T} = 0.8     
&92.77   &\textcolor{blue}{91.91}   &93.06     &94.18   &\textcolor{blue}{99.26}   &94.21  \\

\hline
\end{tabular}
\end{center}
\end{table}
\setlength{\tabcolsep}{1.4pt}

\section{Demonstration of assumption in Sec. 4.5}
\label{append_b}

\begin{figure}[htb]
  \centering
  \includegraphics[width=0.95\linewidth]{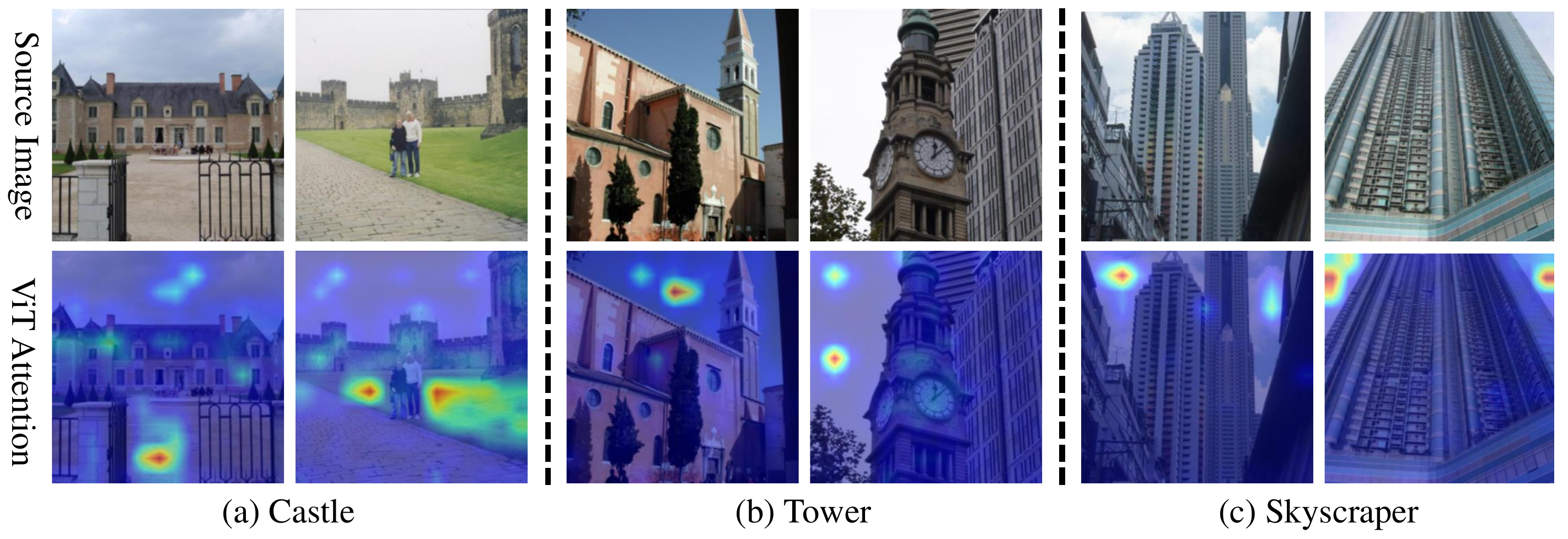}
  \caption{Samples where the representative feature of the image is not easily distinguishable but some shortcut features are easily distinguishable.}
  \label{fig5}
\end{figure}

In Fig.~\ref{fig5}, the first row is the original image, and the second row is the attention map of ViT-S model. 
(a), (b), and (c) are three different categories respectively, and each column is a sample.
As shown in Fig.~\ref{fig5}, without the saliency guidance, the model is prone to focus on regions that are not related to the image categories. 
In addition, since the representative targets of different categories may be similar, such as the left side sample of Fig.~\ref{fig5}(a) and the left side sample of Fig.~\ref{fig5}(b), the right side sample of Fig.~\ref{fig5}(b) and the right side sample of Fig.~\ref{fig5}(c).
It may be difficult to distinguish them directly, but the shortcut features learned by the model make it easier to distinguish these categories instead.
For example, it is possible to distinguish "Castle" from "Tower" by grass and sky, or "Tower” from “Skyscraper” by clean sky and white clouds, or to distinguish different categories just by the position of the sky in the image, etc.
Therefore, we infer that on the FIGRIM~\cite{ZoyaBylinskii2015IntrinsicAE} database, using saliency mask guidance instead has lower performance than not using it for that reason.

\section{Comparison results of using saliency mask on different encoder layers}
\label{append_c}

In the ViT framework, as the depth of encoder layer increases, its attention distance~\cite{dosovitskiy2020image} also changes, so the application of our mask guidance on different layers may also vary.
So we compared the results of applying our mask guidance on different encoder layers, as shown in Tab.~\ref{append_table2} for the mask guidance applied on layers 1, 3, 5, 7 and 9 of ViT-S.
We also compared the results with/without our proposed residual connection separately, as shown in the second column of Tab.~\ref{append_table2}, with $\checked$ representing with residual connection and $\times$ representing without.
Finally, we chose to use mask guidance on the first encoder layer with residual connection.

\setlength{\tabcolsep}{4pt}
\begin{table}[H]
\begin{center}
\caption{Comparison results of using saliency mask on different encoder layers. The Accuracy, AUC, F1-score are reported. \textcolor{red}{Red} and \textcolor{blue}{blue} denote the best and the second-best results, respectively.}
\label{append_table2}
\begin{tabular}{cccccccc}
\hline\noalign{\smallskip}
\multirowcell{2}{Mask\\ Layer} &\multirowcell{2}{Residual\\ Connection} &\multicolumn{3}{c}{INbreast~\cite{InsMoreira2012INbreastTA}} & \multicolumn{3}{c}{FIGRIM~\cite{ZoyaBylinskii2015IntrinsicAE}}\\ \cmidrule(r){3-5} \cmidrule(r){6-8}
&  &Acc. $\uparrow$ & AUC $\uparrow$ & F1 $\uparrow$  & Acc. $\uparrow$ & AUC $\uparrow$ & F1 $\uparrow$  \\
\noalign{\smallskip}
\hline
\noalign{\smallskip}
1   &\checked
&\textcolor{red}{93.98}  &\textcolor{blue}{91.55}  &\textcolor{red}{94.07}      &\textcolor{red}{94.81}  &99.26  &\textcolor{blue}{94.84}    \\
    &$\times$
&90.36  &91.47  &90.37      &94.27  &99.02  &94.42    \\
\midrule
3   &\checked
&\textcolor{blue}{92.77}  &91.13  &93.17      &94.65  &99.23  &94.67    \\
    &$\times$
&89.17  &89.88  &90.03      &94.65  &99.12  &94.60    \\
\midrule
5   &\checked
&91.57  &\textcolor{red}{93.63}  &91.77      &\textcolor{blue}{94.77}  &\textcolor{red}{99.39}  &94.75    \\
    &$\times$
&87.95  &90.39  &88.59      &\textcolor{blue}{94.77}  &99.24  &94.78    \\
\midrule
7   &\checked
&87.95  &88.96  &88.43      &94.28  &\textcolor{blue}{99.35}  &94.29    \\
    &$\times$
&87.95  &90.91  &87.49      &\textcolor{red}{94.81}  &99.21  &\textcolor{red}{94.88}    \\
\midrule
9   &\checked
&91.57  &87.96  &91.07      &94.03  &99.10  &94.02    \\
    &$\times$
&\textcolor{red}{93.98}  &91.22  &\textcolor{blue}{93.84}      &94.13  &99.23  &94.07    \\

\hline
\end{tabular}
\end{center}
\end{table}
\setlength{\tabcolsep}{1.4pt}

\section{Details of dataset and experiment settings}
\label{append_d}

\subsection{Dataset}
\begin{figure}[htb]
  \centering
  \includegraphics[width=0.8\linewidth]{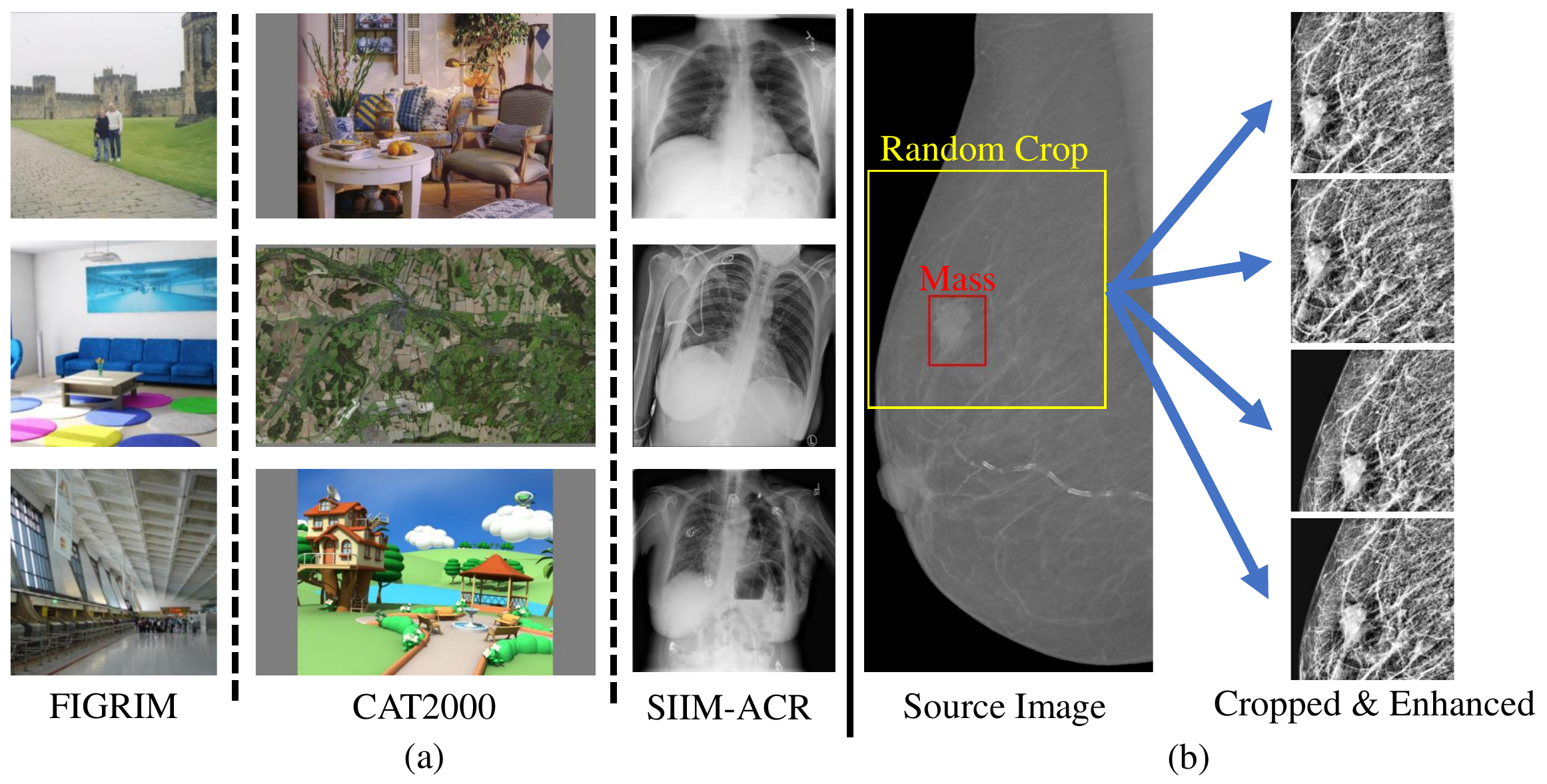}
  \caption{Samples in our experiment. (a) Samples from FIGRIM, CAT2000 and SIIM-ACR. (b) Our data augmentation of images from the INbreast dataset. For the INbreast dataset, we have enhanced the contrast of the images and cropped them randomly.}
  \label{fig6}
\end{figure}

As mentioned in Sec. 4.1, we evaluated our SGT model on INbreast~\cite{InsMoreira2012INbreastTA}, SIIM-ACR~\cite{SIIM-ACR}, FIGRIM~\cite{ZoyaBylinskii2015IntrinsicAE}, and CAT2000~\cite{AliBorji2015CAT2000AL} datasets.
Fig.~\ref{fig6}(a) shows some samples from SIIM-ACR~\cite{SIIM-ACR}, FIGRIM~\cite{ZoyaBylinskii2015IntrinsicAE}, and CAT2000~\cite{AliBorji2015CAT2000AL} datasets.
The FIGRIM~\cite{ZoyaBylinskii2015IntrinsicAE} and CAT2000~\cite{AliBorji2015CAT2000AL} datasets are used to do scene recognition task, and they public the corresponding eye-movement data freely observed by ordinary people.

FIGRIM~\cite{ZoyaBylinskii2015IntrinsicAE} provides eye fixation data for a total of 2787 images spanning 21 indoor and outdoor scene categories.
These images are split into two sets: 630 target images and over 2K filler images.
And we used filler sets as training data and target sets as testing data.

CAT2000~\cite{AliBorji2015CAT2000AL} has 2000 training images and 2000 testing images from 20 different categories with eye tracking data.
We remove 5 categories ('Inverted', 'Jumbled', 'LowResolution', 'Noisy', 'Random') and kept the remaining 15 categories as the final dataset.

The SIIM-ACR~\cite{SIIM-ACR} is a chest X-ray dataset with only pneumothorax disease, and Saab et al.~\cite{KhaledSaab2021ObservationalSF} randomly selected 1,170 images, with 268 cases of pneumothorax and collected gaze data from three radiologists.
We used these 1170 images and randomly split them into 80\% and 20\% as training and testing samples.

For the INbreast~\cite{InsMoreira2012INbreastTA} dataset, due to the small number of samples (410 images) and large image size ($3000 \times 4000$), we performed random cropping and contrast enhancement for each image.
Our task is to classify the images into three categories, no disease, benign masses and malignant masses.
So we take each image as an independent sample.
We first split them into 80\% and 20\% as training and testing samples.
Then, for images with masses (include benign and malignant classes), we only crop the ROI containing the mass area as a sample, as shown in Fig.~\ref{fig6}(b).
And for images without cancer, we cropped randomly on the whole image.
We chose the size of ROI as $1024\times1024$ and enhanced the contrast to every ROI image.
To balance the numbers of training samples in each category, we randomly crop the ROI 8 times for the benign category and 4 times for the malignant category in the training set.
Finally, the training set contains a total of 482 normal samples, 512 benign mass samples, and 472 malignant mass samples. 
In the testing set, we performed a center crop for images with masses and a random crop for the normal category.
All cropped ROI images are resized to $224 \times 224$ pixels finally.

\subsection{Experiment settings}

For all models in our experiment, we use the weights pre-trained on ImageNet~\cite{deng2009imagenet} and fine-tune on each dataset above.
We fine-tune all models, including ResNet~\cite{He_2016_CVPR}, EfficientNet~\cite{tan2019efficientnet}, Swin Transformer~\cite{ZeLiu2021SwinTH}, ViT baselines and our SGT model, using Adam~\cite{kingma2015adam} with $\beta_1=0.9, \beta_2=0.999$, a batch size of 64.
And we train all model with 60 epochs with initial learning rate $1 \times 10^{-4}$ and weight decay $1 \times 10^{-6}$ and 5 epochs of warm-up.
The saliency map is required as input in the training phase of the model, but not in the testing phase.

\subsection{Hardware}
All models were trained on an internal server with 10 NVIDIA GeForce RTX 1080Ti GPUs (11GB).
All experiments used the PyTorch deep learning framework~\cite{paszke2019pytorch}.

\section{Details of our proportion of shortcut learning (PSL) metric}
\label{append_e}

\begin{figure}[htb]
  \centering
  \includegraphics[width=0.8\linewidth]{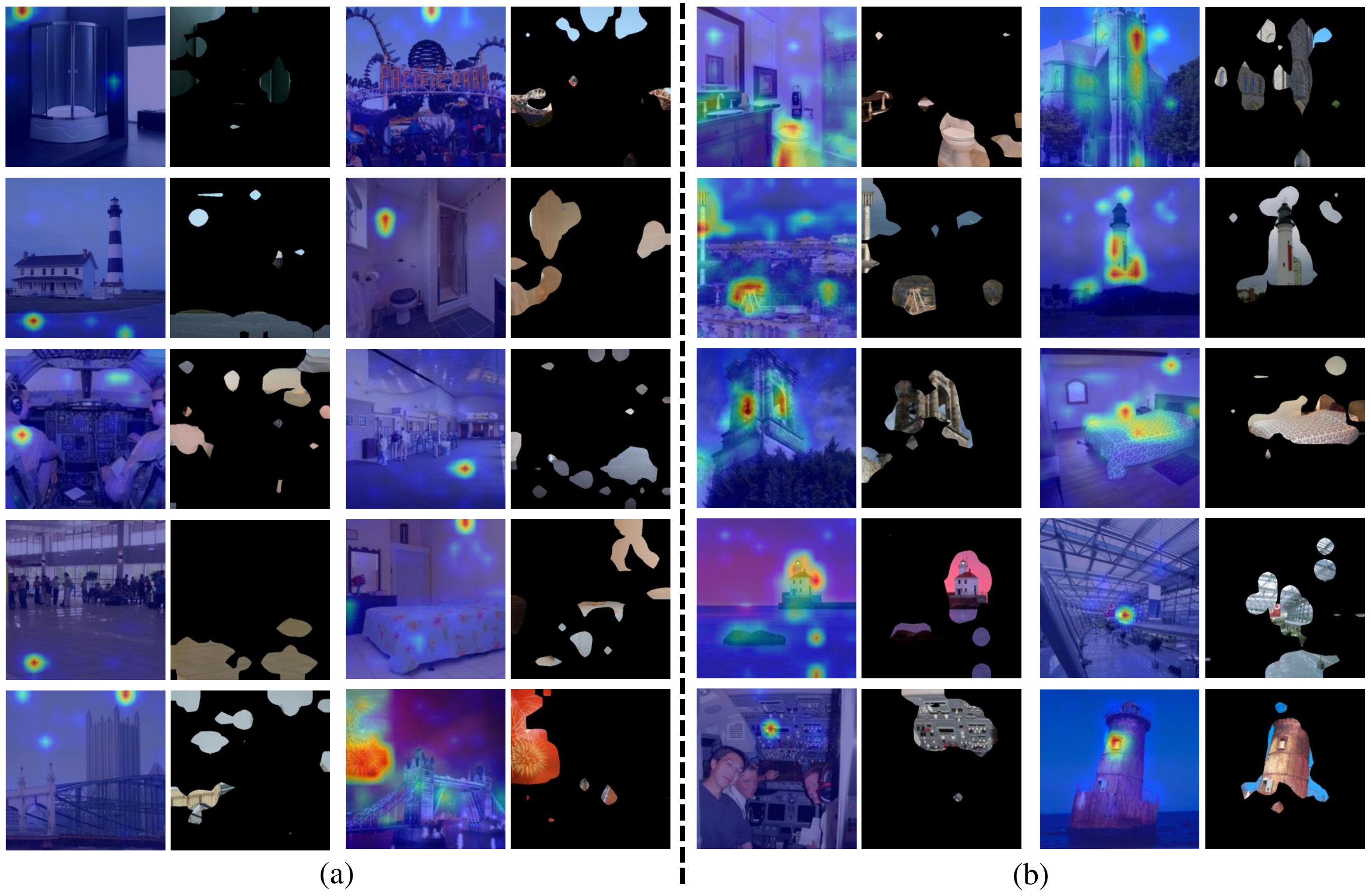}
  \caption{(a) Samples with the shortcut learning phenomenon. (b) Samples without the shortcut learning phenomenon. There are two graphs for each sample, the one on the left is the attention heatmap of the model, and the one on the right is the regional activation map.}
  \label{fig7}
\end{figure}

In our experiment, we introduce a simple method to evaluate the shortcut learning phenomenon.
We first use the Grad-CAM~\cite{RamprasaathRSelvaraju2016GradCAMVE} method to visualize the attention heatmap of models on images.
Then, we check the interest region of the model whether has a significant relationship with the category of image.
As shown in Fig.~\ref{fig7}(a), we consider the existence of shortcut learning if the model focuses on regions unrelated to the image category.
Conversely, as shown in Fig.~\ref{fig7}(b), if the model focuses on regions relevant to the image category, we consider that there is no shortcut learning.
We simply counted the number of images with shortcut learning in the test dataset, and finally obtained a ratio that can directly reflect the severity of the shortcut learning phenomenon, called proportion of shortcut learning (PSL).

\section{Comparison of saliency prediction model and salient object detection model}
\label{append_f}
\begin{figure}[htb]
  \centering
  \includegraphics[width=0.8\linewidth]{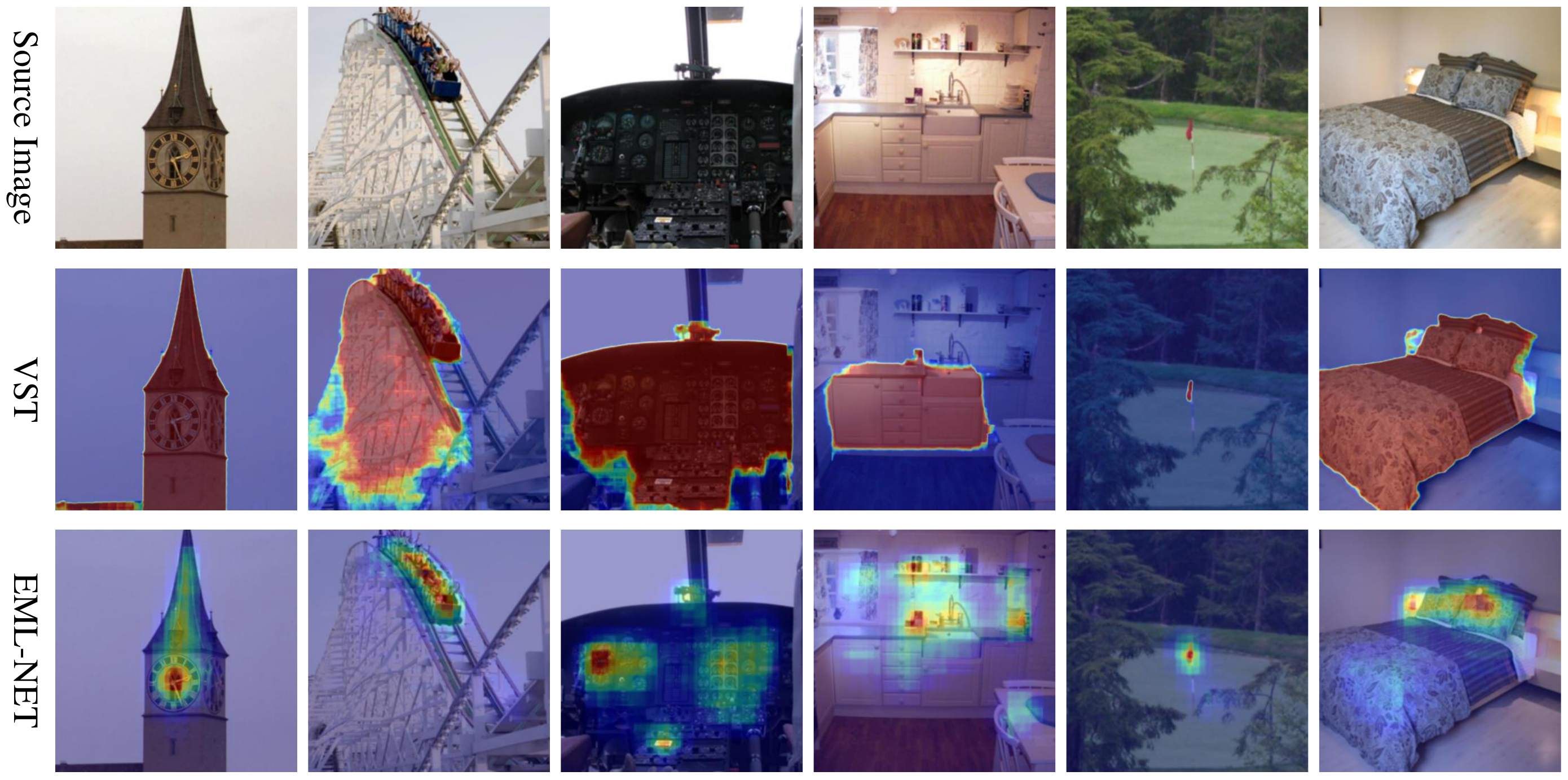}
  \caption{Comparison saliency map generated from a saliency prediction model (EML-NET) and a salient object detection model (VST). The first row is the source image, the second row is the saliency map generated from VST model, and the third row is the saliency map generated from EML-NET model. Each column corresponds to the same example.}
  \label{fig8}
\end{figure}

In our work, we use a saliency map to generate a guidance mask to distill the most discriminative regions for SGT model.
Nowadays, there are two main tasks in artificial saliency: saliency prediction and salient object detection.
The saliency prediction model outputs a Gaussian distributed heatmap for direct simulation of human fixation information.
The salient object detection (SOD) model outputs a mask map, which simulates the human visual perception in locating the most significant object(s) in a scene.
To prove which type of model is more suitable for generating our saliency guidance mask, we chose a saliency prediction model EML-NET~\cite{SenJia2018EMLNETAnEM} and a SOD model VST~\cite{NianLiu2021VisualST} for comparison.
As shown in Fig.~\ref{fig8}, the saliency map generated from VST~\cite{NianLiu2021VisualST} is a mask-type image, which represents some salient regions or objects within the current image as a whole, and the values within the salient regions are all the same.
In contrast, the saliency map generated from EML-NET~\cite{SenJia2018EMLNETAnEM} is a Gaussian distributed heatmap, and the values within the salient regions can be ordered from large to small.
So we can select regions of different importance as a mask according to the ranking of values.
In Sec. 4.5, the ablation experiments on different types of masks also illustrate that the saliency prediction model is more suitable for our task.

\section{More examples of shortcut learning rectification}
\label{append_g}

\begin{figure}[htb]
  \centering
  \includegraphics[width=1.0\linewidth]{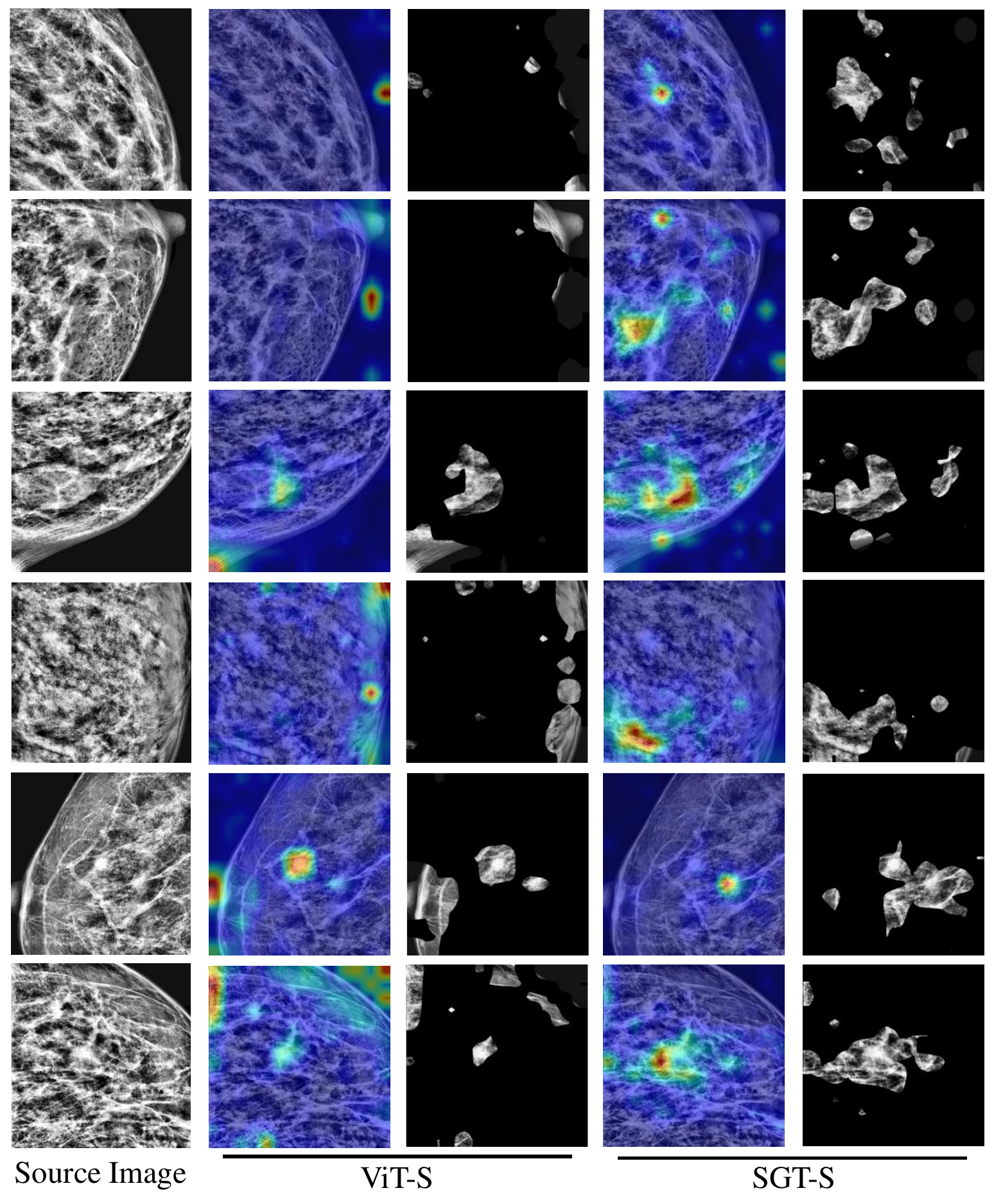}
  \caption{Shortcut learning rectification in INbreast dataset}
  \label{fig9_2}
\end{figure}

\begin{figure}[H]
  \centering
  \includegraphics[width=1.0\linewidth]{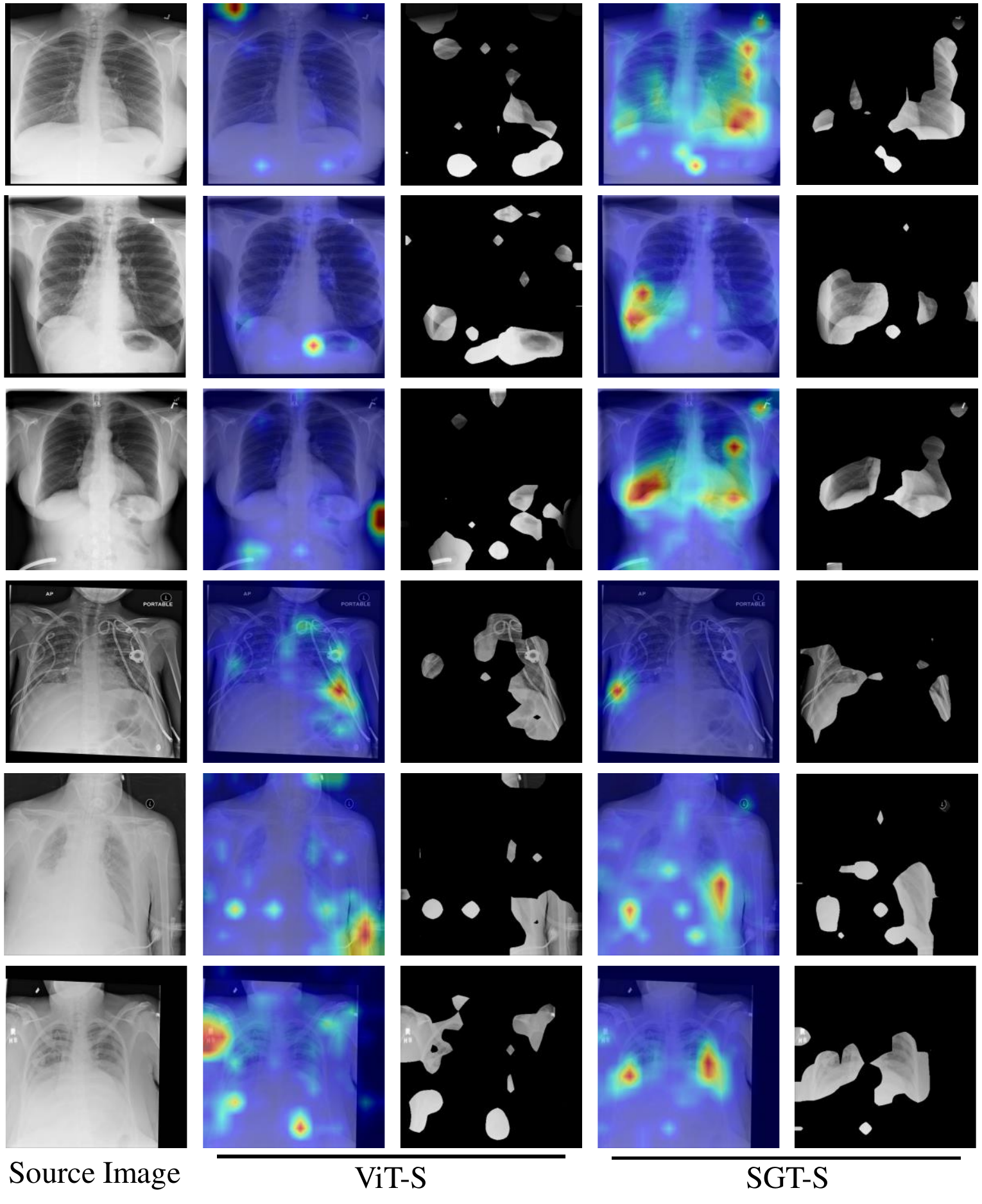}
  \caption{Shortcut learning rectification in SIIM-ACR dataset}
  \label{fig9_1}
\end{figure}

\begin{figure}[H]
  \centering
  \includegraphics[width=1.0\linewidth]{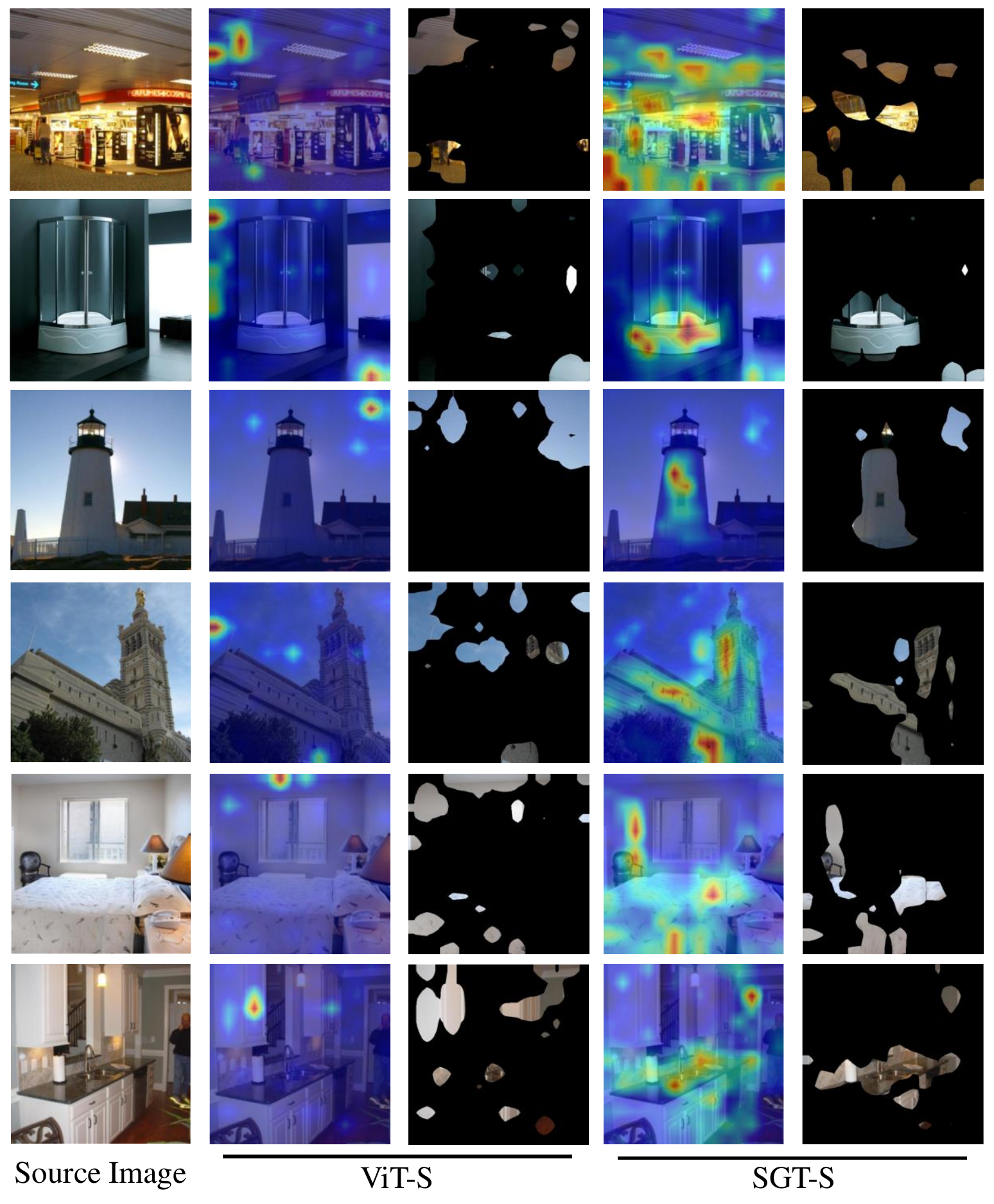}
  \caption{Shortcut learning rectification in FIGRIM dataset}
  \label{fig9_3}
\end{figure}

\begin{figure}[H]
  \centering
  \includegraphics[width=1.0\linewidth]{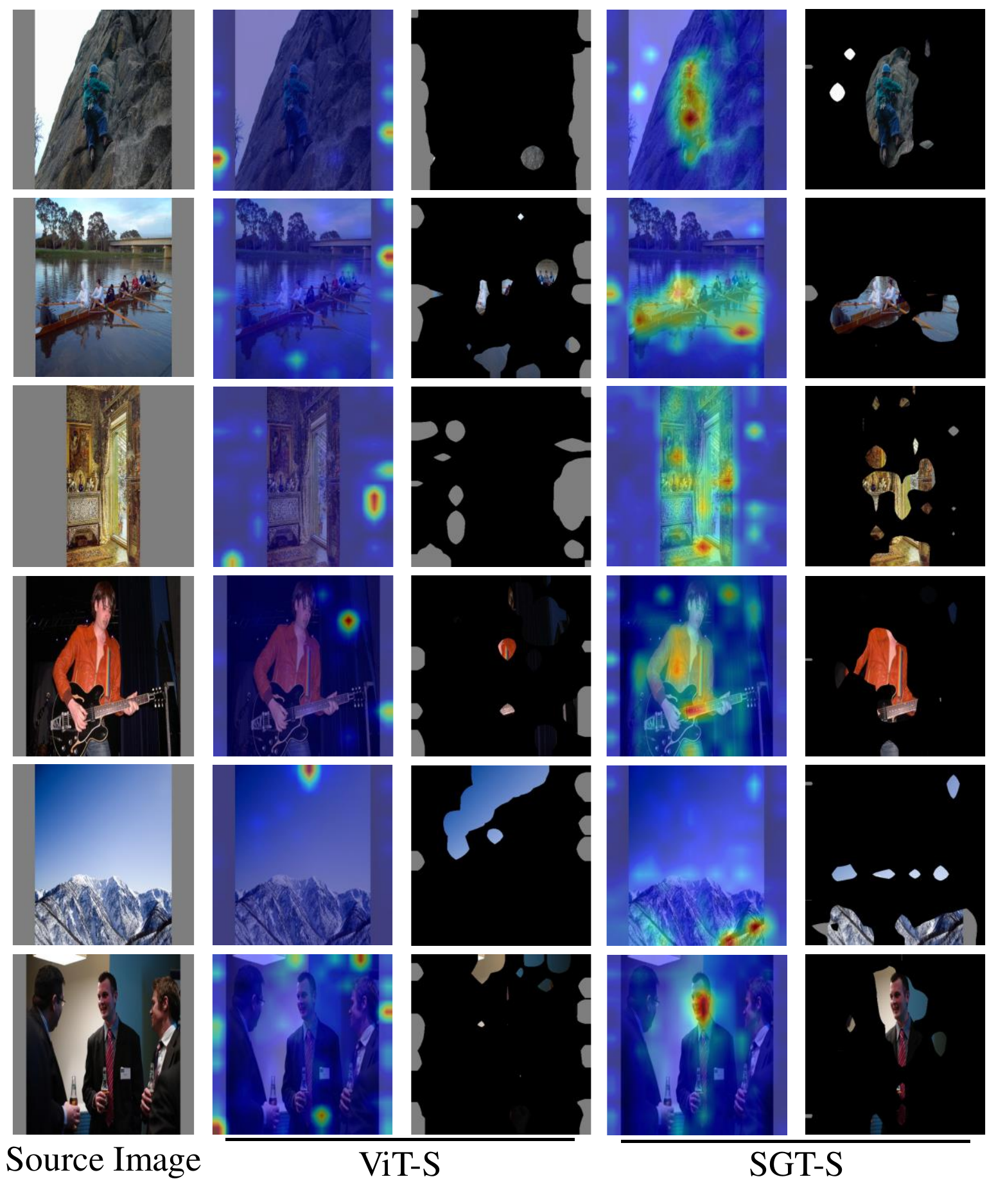}
  \caption{Shortcut learning rectification in CAT2000 dataset}
  \label{fig9_4}
\end{figure}

Fig.~\ref{fig9_2}, Fig.~\ref{fig9_1}, Fig.~\ref{fig9_3} and Fig.~\ref{fig9_4} show more examples of shortcut learning rectification in INbreast~\cite{InsMoreira2012INbreastTA} , SIIM-ACR~\cite{SIIM-ACR}, FIGRIM~\cite{ZoyaBylinskii2015IntrinsicAE} and CAT2000~\cite{AliBorji2015CAT2000AL} datasets.
In these figures, the first column is the source image, the second and third columns are the attention heatmap and activation map of ViT-S model, the fourth and fifth columns are the attention heatmap and activation map of our SGT-S model.
Each row corresponds to the same example.

\section{Eye-tracking system for radiologists}
\label{append_h}
In this section, we introduce a novel solution of eye-tracking system for radiologists.
Most of the datasets currently used to train medical AI models require very clear annotations, such as bounding box or voxel/pixel-level boundaries, which costs much time and effort for radiologists.
However, experienced radiologist's time is precious and costly, which makes the data collection very slow and difficult.
This means that a huge amount of valuable data is not collected and fully exploited. 

To collect high-quality eye gaze data, it is important to take care of the radiologist's experience and not interfere with his/her diagnostic process as much as possible to avoid generating meaningless data or non-subjective diagnosis. 
Nowadays, eye trackers are widely used to collect eye gaze data. 
There are two main types of eye trackers, wearable and screen-based. 
Among them, screen-based eye trackers are commonly used in the medical imaging field because they are less intrusive to radiologists.
However, because of the coordinate mapping problem of eye gaze data, most of the existing works~\cite{karargyris2021creation,stember2019eye,stember2020integrating,wang2022follow,CalvinFNodine2002TimeCO,RaymondBertram2016EyeMO} did not adequately take into account the radiologist's human-computer interaction, and only displayed the image in a fixed form. 
Specifically, the gaze point coordinates collected by the eye tracker are relative to the entire screen, which means that if you want to know exactly where the radiologist is looking at the image you must keep the image positioned fixed on the screen throughout the data collection. 
Then, the screen coordinates of the gaze point can be directly converted to the coordinates on the image.
But in these works, during eye gaze collection, the radiologist can not adjust the size and contrast of the image, or write a diagnosis report, which is not in line with the radiologist's regular work habit and can affect or even reduce the radiologist's diagnostic accuracy.

\begin{figure}[htb]
  \centering
  \includegraphics[width=1.0\linewidth]{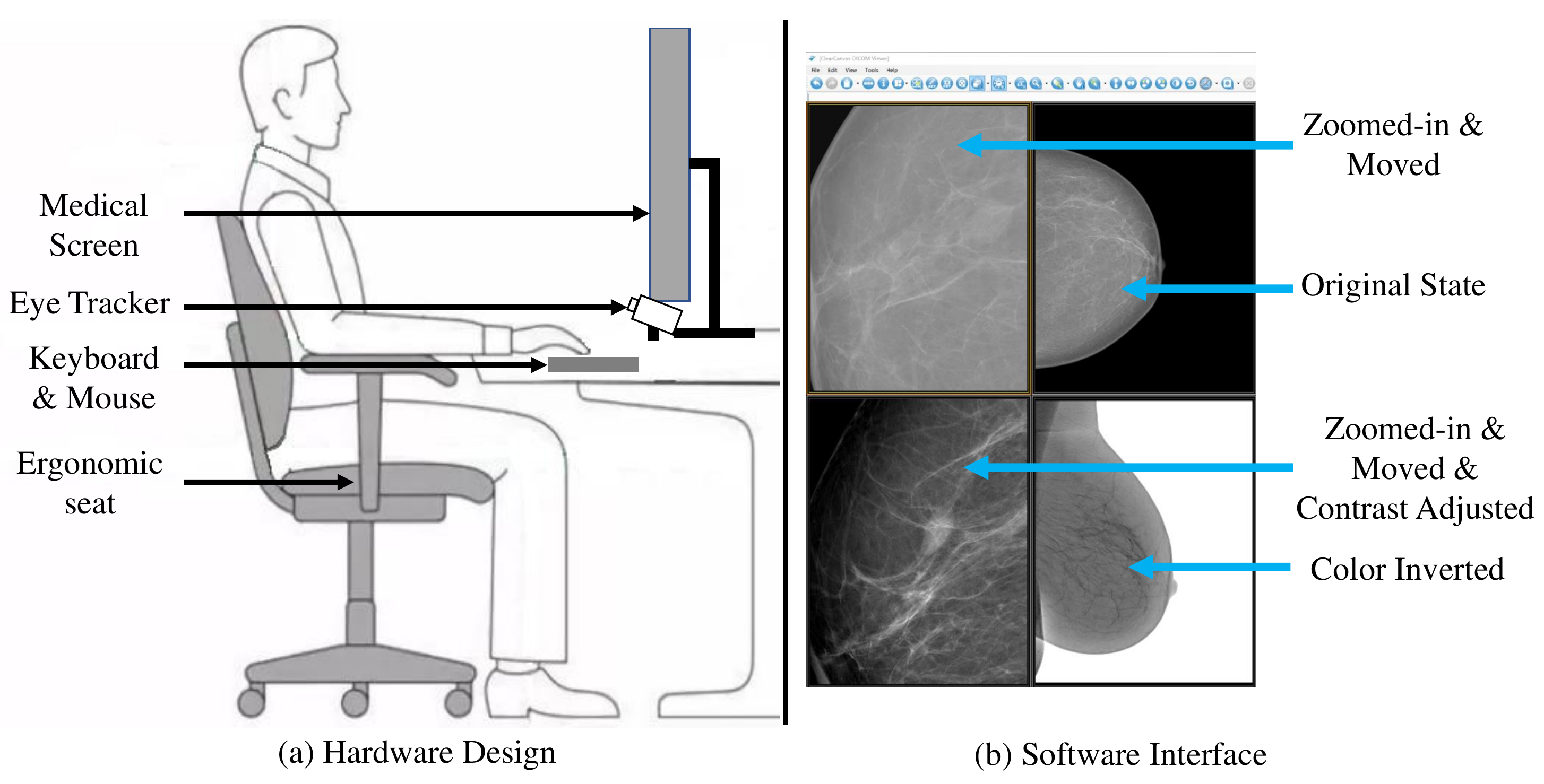}
  \caption{(a) Hardware design of eye-tracking system, (b) Medical image display interface of ClearCanvas}
  \label{fig10}
\end{figure}

To address this problem, we designed a new eye gaze collection system for medical image reading, which maximally restores the diagnostic environment for the radiologists in the regular work. 
Fig.~\ref{fig10}(a) shows our eye-tracking system environment with the radiologist sitting at a desk facing a medical monitor with an eye tracker located below the monitor.
Fig.~\ref{fig10}(b) shows the interface of an open source medical imaging viewer, ClearCanvas~\cite{clearcanvas}.
ClearCanvas supports multi-window viewing, and the radiologist can adjust the image size and position, contrast, etc., which is basically the same as the radiologist's operation in daily work.
In order to solve the coordinate mapping problem, we added four modules to the original software code.
The first is the image status acquisition module.
This module acquires the real-time position of the image on the screen, as well as the status of the zoom ratio, contrast, etc.
The second is the eye-tracking module.
This module is responsible for connecting to the eye-tracker and acquiring the data collected by the eye-tracker.
The third is the diagnosis report module.
we designed a simplified structured report writing module that allows the radiologist to provide a clear diagnosis report after reading.
Last, the coordinate mapping module.
Since the data saved by the first and second modules are with timestamps, coordinate mapping module first synchronizes the image state with the gaze point coordinate according to the timestamp.
At each time point, the image is still relative to the screen, and we can directly convert the gaze point to the image coordinate.
After converting all the data at all time points, we can obtain the eye gaze data for each image.

To sum up, when using our eye-tracking system, radiologists can move the image position and adjust the image size and contrast at well in the process of image reading. 
Meanwhile, they can give a clear diagnosis report using our structured report writing module. 
To our best knowledge, this is the first eye gaze collection system that fully considers the human-computer interaction for radiologists.
We public the code of the above four modules, more details can be found in \url{https://github.com/MoMarky/eye-tracking-system-for-radiologists}.

\end{document}